\documentclass[runningheads]{llncs}
\usepackage{graphicx}
\usepackage{pifont}
\usepackage{tikz}
\usepackage{comment}
\usepackage{amsmath,amssymb} 
\usepackage{color}
\usepackage{multirow}
\usepackage{subfigure}
\usepackage{algorithm} 
\usepackage{algorithmic}
\usepackage{amsmath}  
\usepackage{booktabs}
\usepackage{cleveref}
\usepackage{subfig}
\usepackage{caption}
\usepackage{float}
\newcommand{\eg}{\textit{e}.\textit{g}.}

\usepackage[accsupp]{axessibility}  

\usepackage[width=122mm,left=12mm,paperwidth=146mm,height=193mm,top=12mm,paperheight=217mm]{geometry}

\begin{document}
\pagestyle{headings}
\mainmatter
\def\ECCVSubNumber{2188}  

\title{Plug-and-Play Pseudo Label Correction Network for Unsupervised Person Re-identification} 


\titlerunning{Abbreviated paper title}
%
\author{Tianyi Yan\inst{1,2} \and
Kuan Zhu\inst{1}\and
Haiyun Guo\inst{1}\and
Guibo Zhu\inst{1} \and
Ming Tang\inst{1} \and
Jinqiao Wang\inst{1}
}
\authorrunning{T. Yan et al.}
%
\institute{National Laboratory of Pattern Recognition, Institute of Automation, Chinese Academy of Sciences, Beijing 100190, China \and
School of Artificial Intelligence, University of Chinese Academy of
Sciences, Beijing, 100190, China\\
\email{ \{tianyi.yan, kuan.zhu, haiyun.guo\}@nlpr.ia.ac.cn}
}
\maketitle

\begin{abstract}
Clustering-based methods, which alternate between the generation of pseudo labels and the optimization of the feature extraction network, play a dominant role in both unsupervised learning (USL) and unsupervised domain adaptive (UDA) person re-identification (Re-ID). To alleviate the adverse effect of noisy pseudo labels, the existing methods either abandon unreliable labels or refine the pseudo labels via mutual learning or label propagation. 
However, a great many erroneous labels are still accumulated
because these methods mostly adopt traditional unsupervised clustering algorithms which rely on certain assumptions on data distribution and fail to capture the distribution of complex real-world data. 
In this paper, we propose the plug-and-play \textbf{g}raph-based pseudo \textbf{l}abel \textbf{c}orrection network (GLC) 
to refine the pseudo labels in the manner of supervised clustering. GLC is trained to perceive the varying data distribution at each epoch of the self-training with the supervision of initial pseudo labels generated by any clustering method. It can learn to rectify the initial noisy labels by means of the relationship constraints between samples on the $k$ \textbf{N}earest \textbf{N}eighbor ($k$NN) graph and early-stop training strategy. Specifically, GLC learns to aggregate node features from neighbors and predict whether the nodes should be linked on the graph. Besides, GLC is optimized with `early stop' before the noisy labels are severely memorized to prevent overfitting to noisy pseudo labels. Consequently, GLC improves the quality of pseudo labels though the supervision signals contain some noise, leading to better Re-ID performance.
Extensive experiments in USL and UDA person Re-ID on Market-1501
and MSMT17 show that our method is widely compatible with various clustering-based methods and promotes the state-of-the-art performance consistently.
\keywords{Re-identification, Noise Label Learning, GCN, Clustering}
\end{abstract}

\section{Introduction}
\label{sec:intro}
\begin{figure}
    \subfigure[]{
      \centering
    \includegraphics[width=0.49\linewidth]{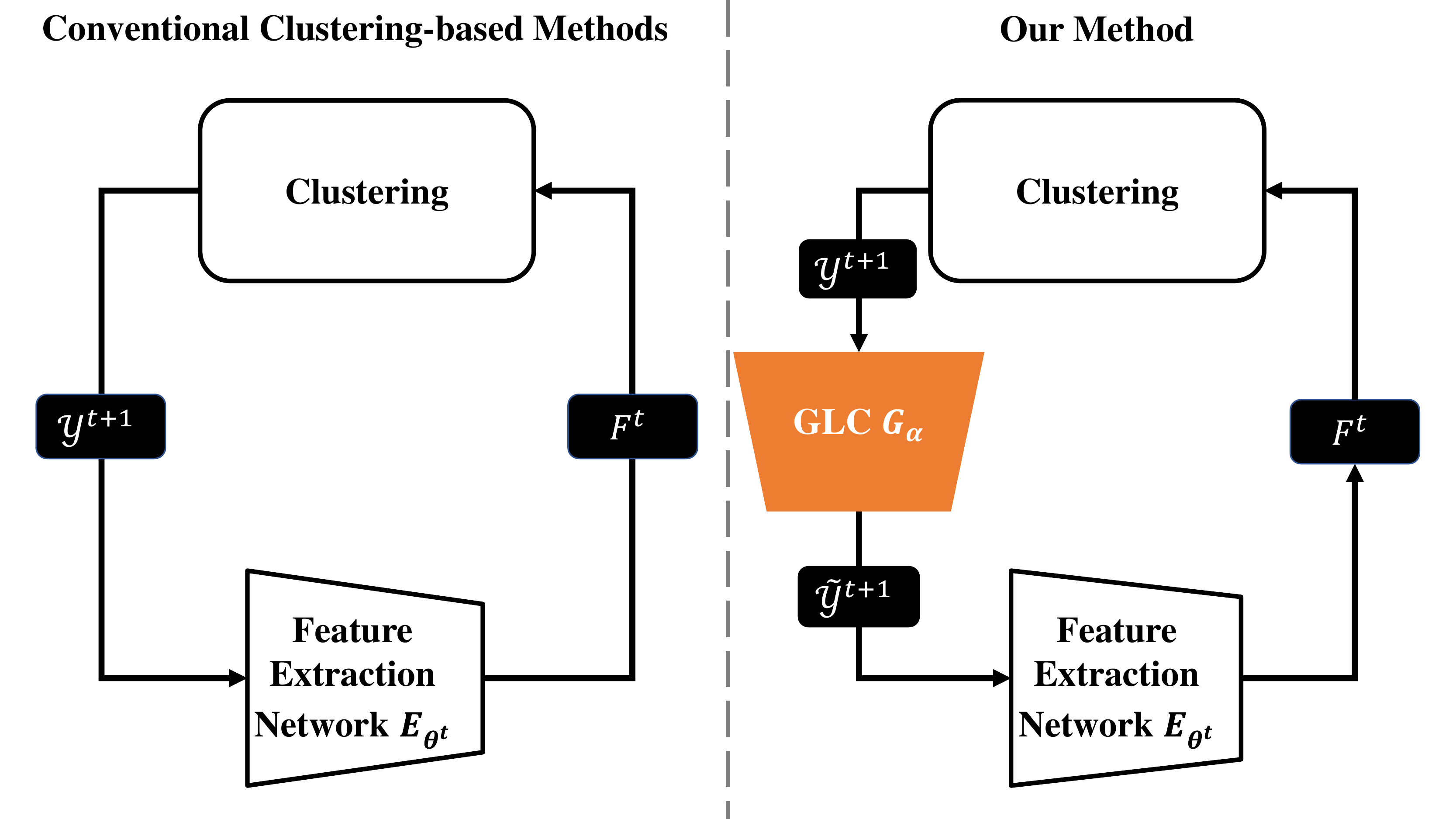}
    }
    \hspace{2em}
    \subfigure[]{
     \centering
  \includegraphics[width=0.45\linewidth]{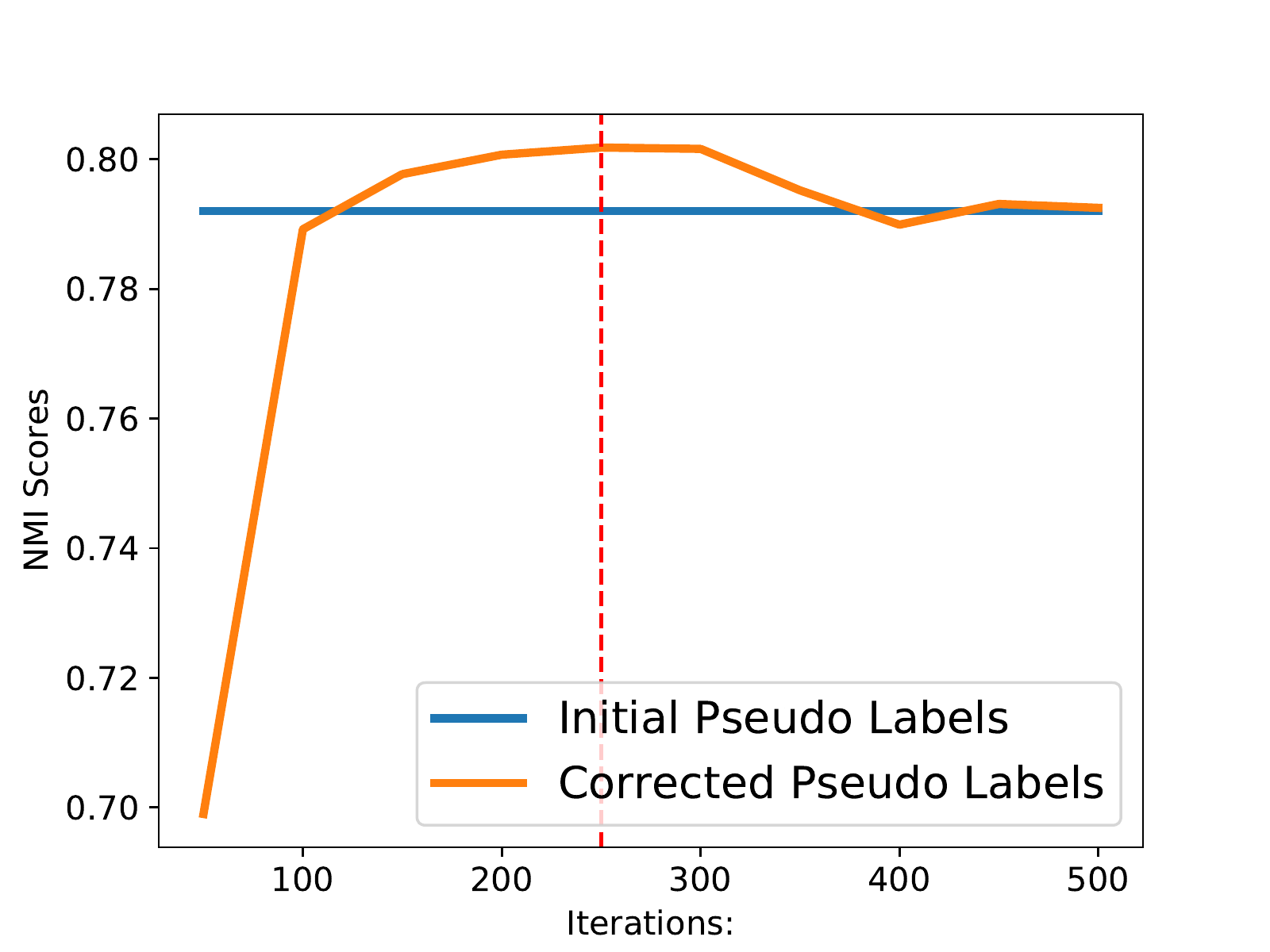}
    }
    \vspace{-1.5em}
    \caption{
    (a) Conventional clustering-based methods alternate between the generation of pseudo labels $y^{t+1}$ via clustering the features $\mathbf{F}^t$, and the optimization of network in $(t+1)$-th epoch. This self-training manner can produce a great many label noises, substantially hindering the training of $E_\theta$. 
    We propose GLC $G_\alpha$ as a post-processing module to refine the initial pseudo labels after each clustering, improving the performance of $E_\theta$. Note that GLC is not required during the testing phase. 
    (b) NMI Scores of pseudo labels at different iterations in the GLC training. We early stop the training of GLC (The dotted line), which prevents the overfitting to the label noise, improving the initial noisy pseudo labels.
    }
    \label{fig:motivation}
\vspace{-1.5em}
\end{figure}

Person re-identification (ReID), which aims to associate person images captured by disjoint cameras, is of great practical value. Recently, unsupervised learning (USL) person ReID \cite{ge2020self,Chen2021ICEIC} and unsupervised domain adaptive (UDA) person ReID \cite{ge2020mutual,wang2021uncertainty,zheng2020exploiting}, which relax the requirement on labeled training data, have received a lot of research interests.
Nowadays, clustering-based approaches \cite{ge2020mutual,isobe2021discriminative,Chen2021ICEIC,zheng2020exploiting,ge2020mutual,zheng2021online}, which alternate between the generation of pseudo labels and the optimization of feature extraction network as shown in \cref{fig:motivation} (a), dominate the community of USL and UDA person ReID.
Although the accuracy of pseudo labels are gradually improved,  
there are inevitable pseudo label noises, which will be accumulated during training, leading to degraded the re-ID performance.

Recently, a series of researches have been presented to reduce the adverse impact of noisy pseudo labels, and can be roughly divided into two categories, reliable pseudo label selection methods \cite{ge2020self,wang2021uncertainty,zhao2020unsupervised,yang2020asymmetric} and  pseudo label refinement methods \cite{ge2020mutual,zhao2020unsupervised}. The former utilize the estimated  reliability of pseudo labels to select credible samples or down-weight unreliable ones to train the feature extractor. The latter achieve label refinement by adopting mutual learning scheme \cite{ge2020mutual}, optimal transport algorithm \cite{zheng2021group} or label propagation \cite{zheng2021online,wu2021mgh}. The existing methods mostly adopt the traditional unsupervised clustering algorithms, such as K-means \cite{lloyd1982least} and DBSCAN \cite{guo2020density}, to generate pseudo labels. These algorithms all rely on certain assumptions on data distribution (\eg,  convex shape, similar size and same density of clusters), failing to capture the distribution of complex real-world data thus generating many erroneous pseudo labels. This motivates us to tackle the noisy pseudo label issue by applying the supervised clustering framework, which can perceive the data distribution from the training samples, and generate better pseudo labels.

In this paper, we propose a plug-and-play graph-based pseudo label correction network (GLC) to improve the accuracy of pseudo labels for USL and UDA person re-ID. 
GLC is trained to capture the data distribution at each epoch with the supervision of pseudo labels generated by any clustering methods. Although the supervision signals contain some noise, GLC can learn to correct the initial noisy pseudo labels by means of the relationship constraint between samples on the graph and the early-stop training strategy. 
GLC is an add-on component to any clustering-based methods which adopt an iterative two-stage training scheme. As shown in fig.1(a), GLC acts as a post-processing module in a plug-and-play way, which is optimized separately from the feature extraction network to rectify current pseudo labels after each clustering. Thus GLC is widely compatible with the existing pseudo label refinement approaches, and adding it can further reduce the remaining pseudo label noise. 

Specifically, we first construct a $k$NN graph with each node denoting a person image and being connected to its k nearest neighbours. Then we formulate the image clustering task as a problem of the link prediction on the graph, by applying several Graph Convolutional Network layers to aggregate node features and a classification layer to predict whether two nodes should be linked under the supervision of pseudo labels. After the GLC training, each node feature is refined by the similarity 
to its neighbors and the supervisory information, possessing more robustness to the initial noise pseudo labels. Besides, considering that deep neural networks first fit the training data with clean labels
\footnote{The clustering-based methods have a basic assumption that most samples from the same person are given the same pseudo label, and in our practice, GLC first fit them.} 
before eventually memorizing the examples with false labels when trained on noisy labels \cite{liu2020early}, we adopt an early-stop strategy during the GLC training to avoid overfitting to the noisy pseudo labels as illustrated in fig.1(b). Finally, in the GLC inference, we conduct link prediction on the whole graph, and the links with low confidence are cut off. Dynamic clusters are generated to fit the current data distribution, and
there are more chances for those false positive image links to be cut off and those false negative image links and outliers to be linked.
To further unleash the potential of GLC, on the one hand, we jointly utilize sample features and ID classification scores to measure the node similarity and link more positive image pairs together during the $k$NN graph construction, considering that classification scores are more robust to the data distribution gap caused by factors like camera variations than raw features \cite{xuan2021intra}.
On the other hand, we re-initialize the parameters of GLC at each epoch to reduce the accumulated errors inherited by the network parameters, and we also re-train the feature extractor from scratch after some epochs to start another self-training process with the corrected pseudo labels, to get rid of the adverse impact of previous noisy pseudo labels. 

Our main contributions are summarized as follows.
(1) We are the first to adopt a neural-network-based supervised clustering framework for USL/UDA person re-ID,  which perceives the data distribution from training samples and adapts better to the dynamic features at each epoch.
(2) We propose the plug-and-play GLC to learn to refine the initial noisy pseudo labels with the relationship constrains between samples on the graph and the early stop training strategy.
(3) Extensive experiments in unsupervised and UDA person ReID on Market-1501 and MSMT17 demonstrate the wide compatibility and consistent performance promotion of our proposed method to various state-of-the-art clustering-based methods.

\section{Related Work}

\textbf{Reliable pseudo label selection.} Pseudo label noises generated by clustering are unavoidable, which substantially hinders the optimization of the feature extraction network. Some approaches try to alleviate this problem by selecting credible samples based on initial clustering results. Co-teaching \cite{yang2020asymmetric} uses two collaborative models that select samples with minimum loss as input to each other. SPCL \cite{ge2020self} considers different clustering thresholds to choose unreliable clusters as outliers. NRMT \cite{zhao2020unsupervised} maintains two networks to make mutual reliable instance selection for training. UNRN \cite{zheng2020exploiting} exploits the uncertainty to re-weight instances in ReID losses calculation, which can also be seen as dropping some samples. 
These methods mitigate the hindrance of noisy pseudo labels by discarding some samples, thus discarding some helpful information simultaneously.
However, traditional unsupervised clustering algorithms still dominate these methods, which fail to capture the distribution of complex real-world data, thus still generating a great many erroneous pseudo labels. Our neural-network-based practice is trained with current training samples, adapting better to the dynamic data distribution in the self-training.
So GLC can re-label both the samples with incorrect labels and outliers, generating better pseudo labels to tackle the problem.

\textbf{Pseudo label refinement.} Recently, researchers propose a series of label refinement works.
MMT \cite{ge2020mutual} trains two collaborative networks under the supervision of off-line refined hard pseudo labels and on-line refined soft pseudo labels.
GLT \cite{zheng2021group} treats the label refinement problem as an optimal transport problem and assigns multi-group pseudo labels to instances.
BUC \cite{lin2019bottom} proposes an online hierarchical clustering framework to generate more reliable pseudo labels. OPLG \cite{zheng2021online} applies dynamic bi-directional cluster adjustment and label propagation to generate better labels. 
MGH \cite{wu2021mgh} introduces metadata (\textit{i}.\textit{e}., Spatio-temporal information) to image similarity calculation and constructs a hypergraph for label refinement. Different from the above methods, we rectify the pseudo labels based on the separately trained network in a supervised clustering way. Aggregating identifiable features enhance the sample feature from its neighborhood, and then the more discriminative feature can be used for label correction. In addition, our GLC is an add-on component to any clustering-based methods to rectify the erroneous pseudo labels.

\textbf{Graph Convolutional Networks.} Recently, Graph Convolutional Network (GCN) \cite{kipf2016semi} is widely applied in computer vision to model the similarity relationship between samples. \cite{zhang2020global} and \cite{zhong2019graph} deploy GCN for label noise cleaning. However, these methods are sub-optimal for person ReID because the category number is fixed other than dynamic, and the noisy labels will be abandoned rather than corrected. 
Besides, some graph neural-network-based clustering methods are proposed \cite{hamilton2017inductive,guo2020density,yang2020learning,yang2020learn,shen2021structure} to learn sample relations from the partial labeled samples. They improve the traditional algorithms like K-means and DBSCAN and achieve the state-of-the-art for features in the stable space. Some researchers apply these graph-based clustering methods to facial images clustering or noise cleaning task. 
Our GLC is inspired by the above methods, but our GLC is trained with noise pseudo labels other than true annotations.
Although the supervisory signals contain some noise, GLC can learn to correct the initial noisy pseudo labels by means of the relationship constrains between samples on the graph as well as the early-stop training strategy.
Moreover, GLC is initialized and trained at each epoch to adapt to the self-training in ReID.

\section{Methodology}

\subsection{Overview}
\begin{figure*}[tbp]
  \centering
   \includegraphics[width=1\linewidth]{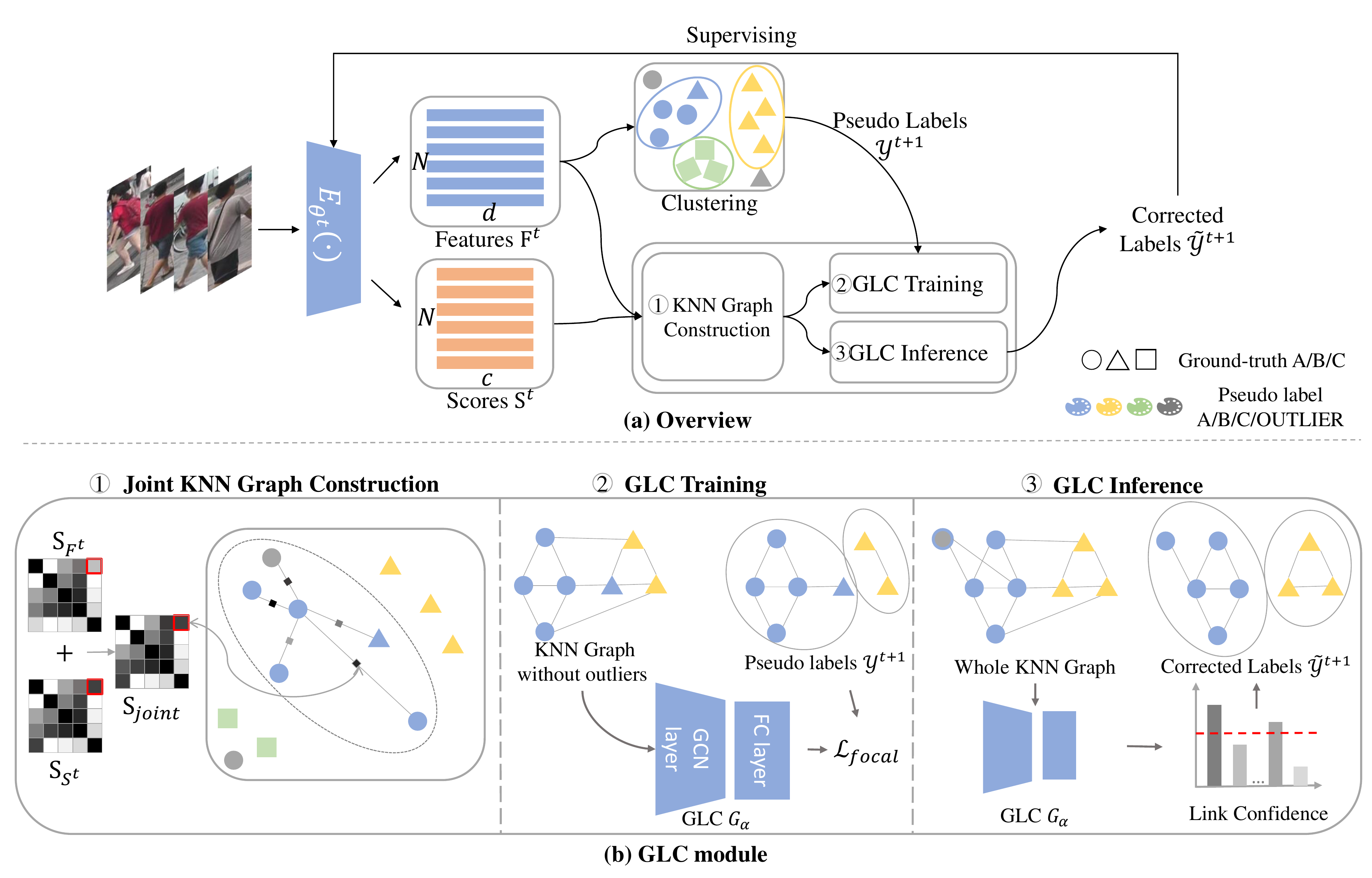}
   \caption{(a) The overview of our method. Our graph-base pseudo label correction network (GLC) 
   can refine the pseudo labels $\mathcal{Y}^{t+1}$ to $\tilde{\mathcal{Y}}^{t+1}$, correcting noisy pseudo labels and relabeling outliers.This way, the accumulation of errors is reduced,
 benefiting the model optimization. Note that our GLC is re-initialized from scratch every epoch.
   (b) Illustration of GLC. \textcircled{1} Features $\mathbf{F}^t$ and classification scores $\mathbf{S}^t$ are used to construct a joint $k$NN graph. \textcircled{2} The GCN and FC layers are trained using $\mathcal{G}$ without outliers and ${\mathcal{Y}}^{t+1}$.
   \textcircled{3} GLC conducts predicting on $\mathcal{G}$ including outliers to determine whether the edges are correct or not, then the corrected labels $\tilde{\mathcal{Y}}^{t+1}$ are obtained by the refined $\mathcal{G}$.}
   \label{fig:graph}
\vspace{-1em}
\end{figure*}

The overview of our method is shown in \cref{fig:graph} (a).
 Let $E_{\theta^t}$ be the feature extraction network with $\theta^t$ as its parameters for the $t$-th epoch. $X=\{x_i | _{i}^N\}$ is the set of images, $N$ is the size of $X$.
 $\mathbf{F}^{t}=\{ \mathbf{f}_i^t = E(x_i|\theta^t)\}$, and $\mathbf{f}_i^t \in \mathbb{R}^d$ is the feature of $x_i$ and extracted by $E_{\theta^t}$. $\mathbf{S}^{t}=\{ \mathbf{s}_{i}^{t} = \mathrm{Softmax}(\mathrm{FC}(\mathbf{f}_i))\} $, where FC is a fully-connected layer and $\mathbf{s}_i\in \mathbb{R}^c$ is the score vector of $\mathbf{f}_i^t$, $c$ is the number of classes.
 Then, through clustering the features $\mathbf{F}^t$, we obtain a set of pseudo labels $\mathcal{Y}^{t+1}=\{y_1,y_2,...,y_N\}$.
 
 Next, $\mathbf{F}^t$ and $\mathbf{S}^t$ of current epoch $t$ are used to construct a joint $k$NN graph ${\mathcal{G}}$. Then, the graph-based pseudo label correction network (GLC) $G_\alpha$ is trained by using the $k$NN graph ${\mathcal{G}}$ and pseudo labels $\mathcal{Y}^{t+1}$, where $\alpha$ is the parameters of GLC. Finally, after the inference process of GLC, the newly corrected pseudo labels $\tilde{\mathcal{Y}}^{t+1}$ are generated to train $E_{\theta^t}$ for the next epoch, achieving $E_{\theta^{t+1}}$.
Note that $G_\alpha$ is re-initialized for each epoch to avoid error accumulation in the network parameters, and also trained with an early-stop strategy to prevent the memorization of the noisy pseudo labels.

\subsection{Graph-based Label Correction Network}
\subsubsection{Joint Similarity Guided Graph Construction}
A robust $k$NN affinity graph ${\mathcal{G}}$ is constructed with $X$ as its nodes, and each node is linked to its $k$ nearest neighbours. The joint similarity matrix is obtained as:
\begin{equation}
  \mathbf{S}_{joint}=\lambda {\mathbf{F}}^\top \mathbf{F}+  (1-\lambda )\mathbf{S}^\top \mathbf{S},
\end{equation}
where $\mathbf{S}_{joint}\in \mathbb{R}^{N\times N}$ and $\lambda$ is the weight to balance the two terms. The epoch $t$ is omitted for simplification. 

We jointly consider the similarity of features ${F}$ and similarity of classification scores $\mathbf{S}$, which can enhance the robustness of similarity to the interference factors. As shown in \cref{fig:graph}(b), the hard positive pairs, which cannot be linked by the common $k$NN graph, can be linked by our $k$NN graph which is constructed by the joint similarity. These added hard positive edges can significantly improve the recall of $k$NN graph.

The sparse symmetric adjacency matrix of ${\mathcal{G}}$ is denoted as $\mathbf{A} \in \mathbb{R}^{N \times N}$ with $A_{i,j}$ as its elements. $A_{i,j}=1$ means the node $i$ is linked with node $j$, and $A_{i,j}=0$ means they are not linked.  The linked samples in ${\mathcal{G}}$ indicate they are in the same clusters. 

It is noted that there are some outliers generated by the clustering algorithm (\eg, DBSCAN) in $X$. In the training stage of GLC, the ourliers are removed from ${\mathcal{G}}$ , while in the inference stage, the outliers are remained in ${\mathcal{G}}$ to be re-labeled.

\subsection{Label Correction with GCN}

After the construction of $k$NN graph ${\mathcal{G}}$, network $G_\alpha$ is trained on the $\mathcal{G}$ without the outliers to adjust the 1 elements of $\mathbf{A}$ as shown in \cref{fig:graph} (b).

The GCN of $G_\alpha$ is designed to learn more robust features. For each node of $\mathcal{G}$, the GCN aggregates its features and its linked neighbors' of the same pseudo label, enhancing their similarity. Following \cite{wang2019linkage}\cite{Li_2019_ICCV}, the modified res-GCN is employed to enhance the features as follows:
\begin{equation}
  \mathbf{H}_{l+1}=\sigma([\mathbf{H}_{l}^{\top},(\hat{\mathbf{A}}\mathbf{H}_{l})^{\top}]^{\top}\mathbf{W}_{l}),
\end{equation}
where $\hat{\mathbf{A}}=\mathbf{D}^{-1}(\mathbf{A}+\mathbf{I})\mathbf{D}$, $\mathbf{D}$ is a diagonal degree matrix. $\mathbf{A}$ is the adjacency matrix of graph $\mathcal{G}$. $\mathbf{W}_{l}$ is the learnable matrix. $\mathbf{H}_{l}$ represents the feature output by $l$-th layer and $\mathbf{H}_0$ is $\mathbf{F}^t$. $\sigma$ is the activation function ReLU\cite{relu}.

Next, a binary classifier is adopted to predict the confidence of the links in $k$NN graph based on the enhanced features. Its supervision is obtained from the pseudo labels $\mathcal{Y}^{t+1}$. The edges whose nodes are of the same pseudo label are positive samples and those whose nodes are of different pseudo labels are negative ones. Thus, the labels of edges are obtained as follows:
\begin{equation}
  y^{e}_{i}=\begin{cases}
  1, y_{v^{i}_1}=y_{v^{i}_2}
 \\0,y_{v^{i}_1}\ne y_{v^{i}_2}
\end{cases}
\end{equation}
where $i$ is the index of the $i$-th edge, $v^{i}$ is the index of the two nodes connected by the $i$-th edge.
Considering the imbalance between positive and negative edges on the $k$NN graph, we adopt Focal loss \cite{focal} to focus on the hard positive samples. The loss is calculated by:
\begin{equation}
  L=\frac{1}{N}(\sum_{y^{e}_i=1}^{m}-(1-p^e)^\gamma log(p^e )+\sum_{y^{e}_i=0}^{n}-{p^e}^\gamma log(1-p^e ) ) 
\end{equation}
where $p^e$ is the prediction probability of the $i$-th edge, m is the number of positive samples and n is the number of negative samples, $N=m+n$.

\begin{figure}[htbp]
  \centering
  \includegraphics[width=0.9\linewidth]{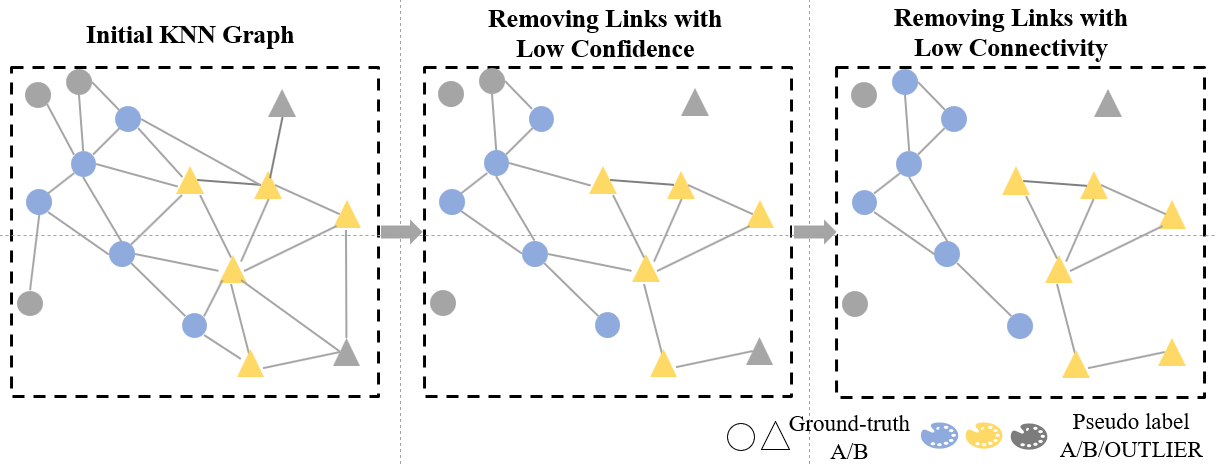}
  \caption{Illustration of removing unreliable edges. Confidences of the edges are obtained by GLC and edges with low confidence are removed. Besides, edges with low $NC$ are also removed. Finally, the linked nodes are assigned with the same pseudo label.}
  \label{fig:inference}
\vspace{-1em}
\end{figure}

After training, $G_\alpha$ has the capability to aggregate features of nodes to enhance their features and predict whether the edges among nodes are correct.
In the inferring stage, $G_\alpha$ takes the $\mathcal{G}$ including the outliers as input, and produces the confidence of edges as shown in \cref{fig:graph} (b).
The edges with lower confidence than threshold $\tau_1$ are removed while others are reserved. We also refine edges by node connectivity (NC) to further remove hard negative edges as shown in Figure~\ref{fig:inference}.
NC of nodes $i$ and $j$ is expressed as:
\begin{equation}
  NC_{i,j}=max(\frac{N_{share}}{Ni},\frac{N_{share}}{Nj})
\end{equation}
where $N_{share}$ is the number of edges connected to the same nodes by node $i$ and node $j$, $N_i$ and $N_j$ are the numbers of edges connected to nodes $i$ and $j$, respectively. 
Another threshold $\tau_2$ is used to remove the link whose nodes have low node connectivity. The parameter analysis of $\tau_1$ and $\tau_2$ can be referred to supplementary materials.

Finally, we obtain a refined $k$NN graph. The nodes that are connected directly or indirectly are assigned with the same pseudo label. After the whole label correction process, noisy pseudo labels are corrected and outliers are re-labeled.
All new pseudo labels are employed to train feature extraction network for the next epoch.
This way, the accumulation of errors is reduced in the self-training process, benefiting the model optimization.

\subsection{Early stop and Restart Training Strategy}
Considering that deep neural networks firstly fit the training data with clean labels before eventually memorizing the examples with false labels when trained on noisy labels \cite{liu2020early}, we early stop our GLC training at $t_e$ to avoid overfitting to the noise pseudo labels, leading to more accurate pseudo labels. 
Inspired by \cite{cascante2020curriculum}, which re-initializes the model parameters each epoch to solve the concept drift problem in the semi-supervised task, we design our re-initialization training strategy for USL and UDA person ReID. 
On the one hand, we re-initialize the parameters of graph-based label correction network every epoch. On the other hand, after a certain number of epochs, we re-initialize the feature extraction network from scratch to start another self-training process with the last corrected pseudo labels. These can further remove the accumulated errors. 

\section{Experiments}
\subsection{Datasets}
We evaluate our method on two widely used person ReID datasets, including Market-1501 \cite{market} MSMT17 \cite{msmt}. Market-1501 consists of 32,668 images of 1,501 identities shot by 6 cameras, of which 12,936 images of 751 identities are used for training and 19,732 images of 750 identities are for testing.
MSMT17 is the largest and most challenging dataset. It consists of 126,441 bounding boxes of 4,101 identities shot by 15 cameras, in which 32,621 images of 1,041 identities are used for training. For performance evaluation of person ReID methods, we adopt mean Average Precision (mAP) and the CMC-$k$ (a.k.a, Rank-$k$ matching accuracy), which represents the probability that a correct match appears in the top-$k$ ranked retrieved results.

\subsection{Implementation Details}
We implement our method by adding GLC and restart strategy into the existing clustering-based unsupervised and UDA person ReID methods, \textbf{i}.\textbf{e}. Sbase \cite{zheng2020exploiting}, IDM \cite{dai2021idm} for UDA setting and CAP \cite{wang2021cameraaware}, ICE \cite{Chen2021ICEIC} for unsupervised setting.
Follow the practice of most graph-based clustering methods \cite{yang2020learn}\cite{shen2021structure}, we set $k$=50 for $k$NN graph construction. We implement GLC with one GCN layer and one FC layer. In terms of the restart strategy, we only re-initialize the feature extraction network for once during training. The SGD optimizer is adopted to optimize the GLC with weight decay set to 1e-5 and initial learning rate set to 0.1. As for inference stage, the threshold $\tau_1$ and $\tau_2$ to cut off the edges are both set to 0.6. Our experiments are conducted on 4 Nvidia 3090 GPUs.
\subsection{Comparison with the State-of-the-art Methods}
\textbf{Unsupervised person ReID.} CAP \cite{wang2021cameraaware} is one of the most popular unsupervised person ReID methods which introduce camera-aware proxies to deal with large intra-ID variance caused by different cameras and generate more reliable pseudo labels. ICE \cite{Chen2021ICEIC} is the latest state-of-the-art unsupervised person ReID method which leverages inter-instance pairwise similarity scores to boost previous class-level contrastive ReID methods. We add GLC and restart training strategy gradually into CAP and ICE to evaluate the effectivenss of our method and compare them with the state of the arts in \cref{tab:state-of-the-art_u}. Adding GLC and restart strategy brings a consistent performance improvement on all the datasets with either CAP or ICE. To be specific, GLC improves the mAP by 0.9\% and 1.3\% on Market1501, MSMT17 respectively, demonstrating its effectiveness for pseudo label refinement and benefits for the training of the feature extraction network. By further adding restart strategy to feature extraction network, the ReID performance continues to increase by 0.8\% and 0.8\% at mAP on Market1501, MSMT17 respectively. This indicates that the feature extraction network can be better optimized by forgetting the historical mistakes inherited by network parameters via re-initialization. Finally, adding our method to ICE increases the state-of-the-art performance by a significant margin of 1.7\% and 2.1\% at mAP on Market1501, MSMT17 respectively.
\begin{table*}[h]
  \centering
  \caption{Comparisons to the state-of-the-arts on multiple benchmarks for unsupervised person ReID. $^*$ means the re-implementation by this paper with the official released codes. "Ours$^-$" denotes only adding GLC to the baseline methods while "Ours$^+$" indicates adding both GLC and restart training strategy. "-" denotes the results are not reported. The top results are highlighted in bold.}
    \begin{tabular}{c|c|cccc|cccc}
    \toprule
    \multirow{2}{*}{Method} & \multirow{2}{*}{Reference} & \multicolumn{4}{c}{Market1501} & \multicolumn{4}{c}{MSMT17} \\
\cline{3-10}          &       & mAP   & R1    & R5    & R10   & mAP   & R1    & R5    & R10 \\
    \hline
    MMCL  & CVPR20 & 45.5  & 80.3  & 89.4  & 92.3  & 49.8  & 11.2  & 35.4  & 44,8 \\
    HCT   & CVPR20 & 56.4  & 80    & 91.6  & 95.2  & —     & —     & —     & — \\
    SpCL  & NeurIPS20 & 72.6  & 87.7  & 95.2  & 96.9  & 19.1  & 42.3  & 55.6  & 61.2 \\
    OPLG  & ICCV21 & 78.1  & 91.1  & 96.4  & 97.7  & 26.9  & 53.7  & 65.3  & 70.2 \\
    CAP   & AAAI21 & 79.2  & 91.4  & 96.3  & 97.7  & 36.9  & 67.4  & 78    & 81.4 \\
    ICE   & ICCV21 & 82.3  & 93.8  & 97.6  & 98.4  & 38.9  & 70.2  & 80.5  & 84.4 \\
    \hline
    CAP*  & AAAI21 & 77.9  & 90.5  & 95.5  & 97.1  & 36.5  & 67.5  & 77.6  & 81.4 \\
    CAP+Ours$^-$ & this paper & 78.8  & 91.1  & 95.6  & 97.2  & 37.6  & 69.2  & 79.2  & 82.6 \\
    CAP+Ours$^+$ & this paper & 80.1  & 91.4  & 96.5  & 98.1  & 38.5  & 70.5  & 81    & 83.9 \\
    ICE*  & ICCV21 & 81.8  & 92.8  & 97.5  & 98.2  & 38.8  & 71    & 81.1  & 84.3 \\
    ICE*+Ours$^-$ & this paper & 82.7  & 93.6  & 97.8  & \textbf{98.6} & 40.1  & 72    & 81.9  & 85.2 \\
    ICE*+Ours$^+$ & this paper & \textbf{83.5} & \textbf{93.9} & \textbf{97.9} & \textbf{98.6} & \textbf{40.9} & \textbf{72.4} & \textbf{82.3} & \textbf{85.9} \\
    \bottomrule
    \end{tabular}%
  \label{tab:state-of-the-art_u}%
\vspace{-1em}
\end{table*}%

\textbf{Unsupervised domain adaptive person ReID.} Sbase \cite{zheng2020exploiting} is a classical UDA baseline which is based on mean teacher framework and utilizes cross batch memory bank.
IDM \cite{dai2021idm} utilizes the characteristics of intermediate domains as the bridge to better transfer the source knowledge to the target domain.
We gradually add GLC and restart strategy into these two methods and compare our methods with other state-of-the-art UDA person ReID methods on multiple benchmarks. The results are illustrated in \cref{tab:state-of-the-art_uda}. As we can see, the adding of GLC and restart strategy also brings consistent performance improvement for UDA person ReID.  
With our GLC and restart strategy, the performance of IDM is promoted to be the best on both benchmarks. The total increase at mAP for IDM is 1.7\% and 4.0\% on both benchmarks respectively. 
It's worth noting that both CAP and Sbase are proposed to deal with the adverse effect of noisy pseudo labels. The results in \cref{tab:state-of-the-art_u} and \cref{tab:state-of-the-art_uda} demonstrate that, 
with GLC and restart strategy to further reduce the accumulated errors, our method can promote their performance significantly. That is to say, our method is widely compatible with the existing clustering-based methods, whether they have refined the pseudo labels or not.

\begin{table*}[h]
  \centering
  \caption{Comparisons to the state-of-the-arts on multiple benchmarks for unsupervised domain adaptive person ReID.}
    \begin{tabular}{c|c|ccccc|ccccc}
    \toprule
        \multirow{2}{*}{Method} & \multirow{2}{*}{Reference} & \multicolumn{5}{c}{Market1501}        & \multicolumn{5}{c}{MSMT17} \\
\cline{3-12}          &       & source & mAP   & R1    & R5    & R10   & source & mAP   & R1    & R5    & R10 \\
\hline
    SSG   & ICCV19 & MSMT & -    & -    & -    & -    & Market & 13.2  & 31.6  & -    & 49.6 \\
    MMT   & ICLR20 & MSMT & -    & -    & -    & -    & Market & 22.9  & 49.2  & 63.1  & 68.8 \\
    SpCL  & NeurIPS20 & MSMT & 77.5  & 89.7  & 96.1  & 97.6  & Market &       &       &       &  \\
    UNRN  & AAAI21 & MSMT & -    & -    & -    & -    & Market & 25.3  & 52.4  & 64.7  & 69.7 \\
    GLT   & CVPR21 & MSMT & -    & -    & -    & -    & Market & 26.5  & 56.6  & 67.5  & 72.0 \\
    OPLG  & ICCV21 & MSMT & 80.2  & 91.4  & -    & -    & Market & 28.4  & 54.9  & -    & - \\
    P$^2$LR & AAAI22 & MSMT & -    & -    & -    & -    & Market & 29.0  & 58.8 & 71.2 & 76.0 \\
    TDRL  & ICCV21 & MSMT & -    & -    & -    & -    & Market & 35.8  & 65.8  & -    & - \\
IDM   & ICCV21 & MSMT & -    & -    & -    & -    & Market & 33.5  & 61.3  & 73.9  & 78.4 \\
\hline    
    Sbase* & AAAI21 & MSMT & 78.9  & 91.5  & 95.8  & 97.5  & Market & 25.3  & 54.4  & 65.8  & 70.8 \\
    Sbase+Ours- & this paper & MSMT & 80.7  & 93.5  & 97.7  & 98.5  & Market & 26.8  & 56.2  & 67.3  & 71.4 \\
    Sbase+Ours+ & this paper & MSMT & 81.4  & 93.4  & 97.7  & 98.5  & Market & 28.4  & 56.7  & 68.2  & 72.6 \\
    IDM*  & AAAI21 & MSMT & 82.9  & 92.5  & 97.2    & 98.5  & Market & 34.6  & 62.9  & 74.8  & 79.7 \\
    IDM+Ours- & this paper & MSMT & 83.9  & 93.2  & 97.5  & 98.4  & Market &37.8   &66.0   &76.8   & 80.9  \\
    IDM+Ours+ & this paper & MSMT & \textbf{84.6} & \textbf{94.2} & \textbf{98.0} & \textbf{98.8} & Market &\textbf{41.1} & \textbf{68.7} & \textbf{80.3}  & \textbf{84}\\
    \bottomrule
    \end{tabular}%
  \label{tab:state-of-the-art_uda}%
\vspace{-1em}
\end{table*}%

\subsection{Ablation Studies}
In this section, we evaluate the effectiveness of each component in our approach with both unsupervised and UDA person ReID methods by conducting ablation studies on Market1501 and MSMT17. The results are shown in \cref{tab:ablation_u} and \cref{tab:ablation_uda}. \textbf{GLC$^-$} denotes only using raw features to measure sample similarity in kNN graph construction. \textbf{RS} denotes restart training strategy. \textbf{M} denotes Market-1501, \textbf{MS} denotes MSMT17.
\begin{table}[]
\centering
  \begin{minipage}{0.48\textwidth}
  \centering
    \caption{The effectiveness of each component of our method on multiple benchmarks for UDA ReID.}
    \begin{tabular}{c|cc|cc}
    \toprule
    \multirow{2}{*}{Method} & \multicolumn{2}{c}{M to MS} & \multicolumn{2}{c}{MS to M} \\
\cline{2-5} & mAP   & R1   & mAP   & R1 \\
    \hline
    Sbase* & 25.3  & 54.4 & 78.9  & 91.5 \\
    +GLC$^-$ & 26.6  & 58.7  & 80.5  & 93.0 \\
    +GLC  & 26.8  & 56.2    & 80.7  & 93.5 \\
    +RS & 26.4  & 55.7 & 80.2  & 92.1 \\
    +GLC\&RS & 28.4    & 56.7  & 81.4  & 93.4 \\
    \bottomrule
    \end{tabular}
  \label{tab:ablation_uda}%
\end{minipage}\hspace{1em}\begin{minipage}{0.48\textwidth}
\centering
    \caption{The effectiveness of each component of our method on multiple benchmarks for USL person ReID.}
    \vspace{1em}
    \begin{tabular}{c|cc|cc}
    \hline
    \multirow{2}{*}{Method} & \multicolumn{2}{c}{M} & \multicolumn{2}{c}{MS} \\
    \cline{2-5}  & mAP   & R1    & mAP   & R1 \\
    \hline
    CAP*   & 77.9  & 90.5  & 36.5  & 67.5 \\
    +RS & 78.5  & 90.9  & 37.8  & 68.9 \\
    +GLC  & 78.8  & 91.1  & 37.6  & 69.2 \\
    +GLC\&RS & 80.1  & 91.4  & 38.5  & 70.5 \\
    \bottomrule
    \end{tabular}%
  \label{tab:ablation_u}%
\end{minipage}
\vspace{-1em}
\end{table}

\begin{figure}[h]
  \centering
  \subfigure[Sbase* and Sbase* + GLC.]{
    \centering
   \includegraphics[width=0.4\linewidth]{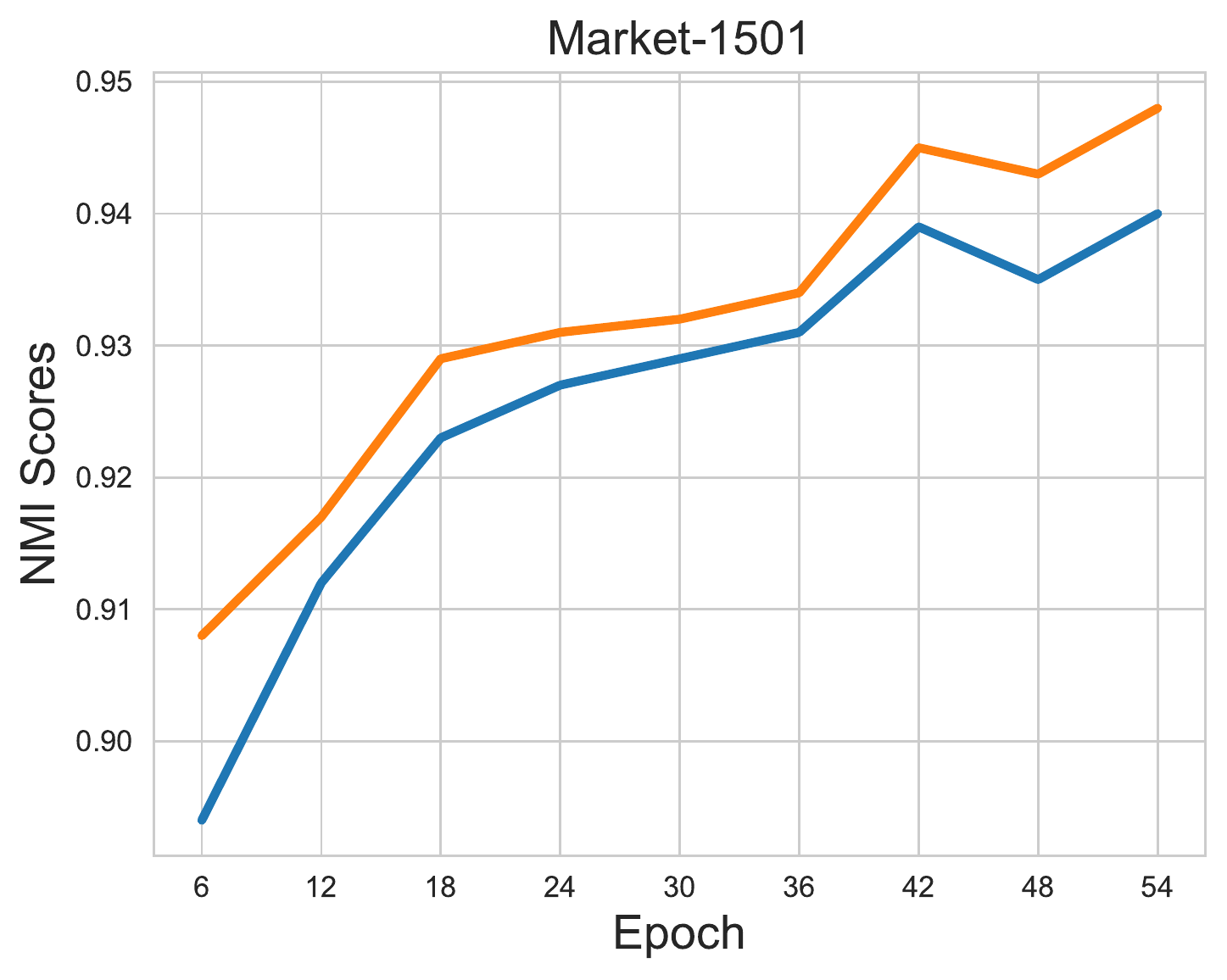}
  }
  \subfigure[CAP* and CAP* + GLC.]{
    \centering
   \includegraphics[width=0.4\linewidth]{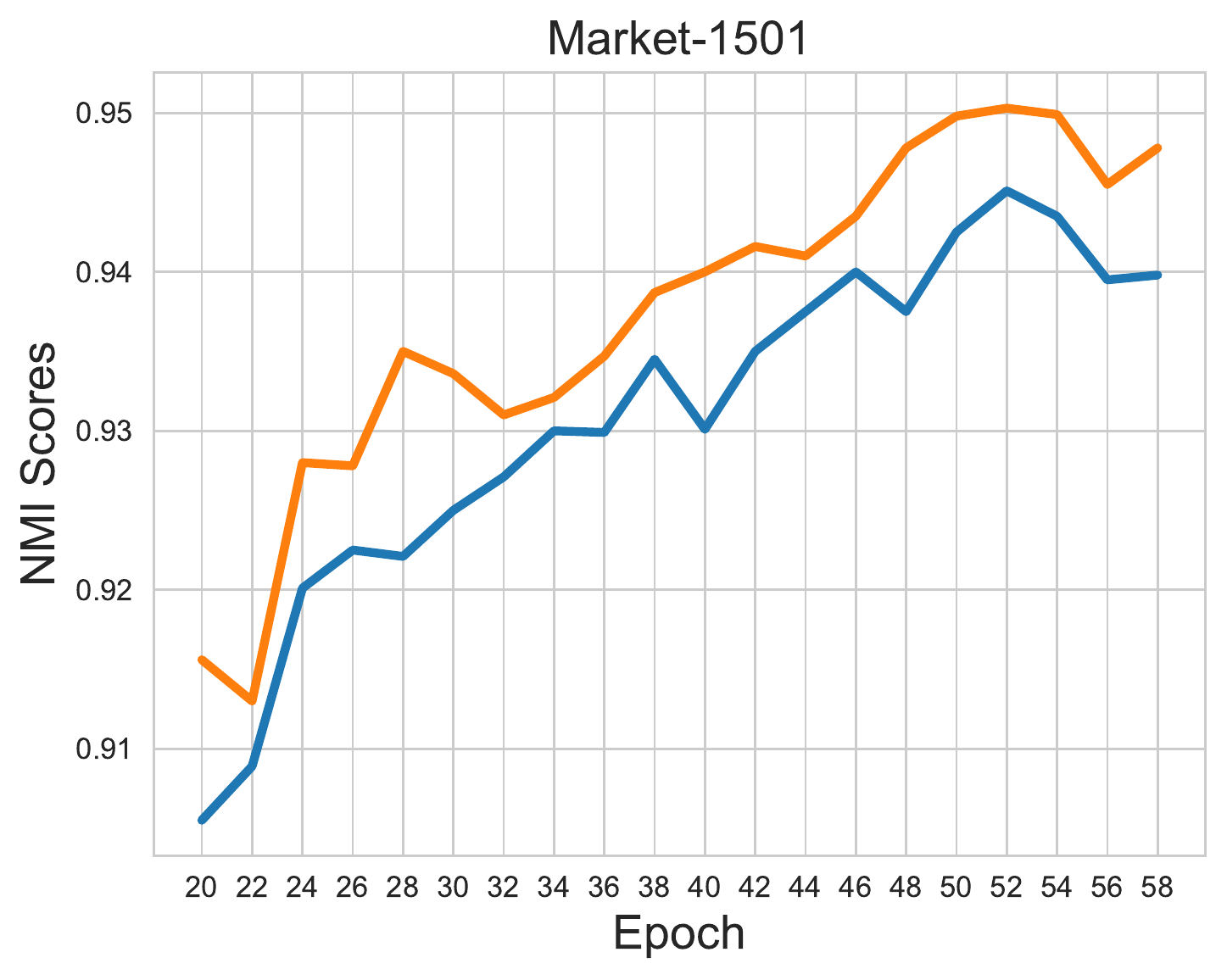}
  }

  \subfigure[Sbase* and Sbase* + GLC.]{
    \centering
  \includegraphics[width=0.4\linewidth]{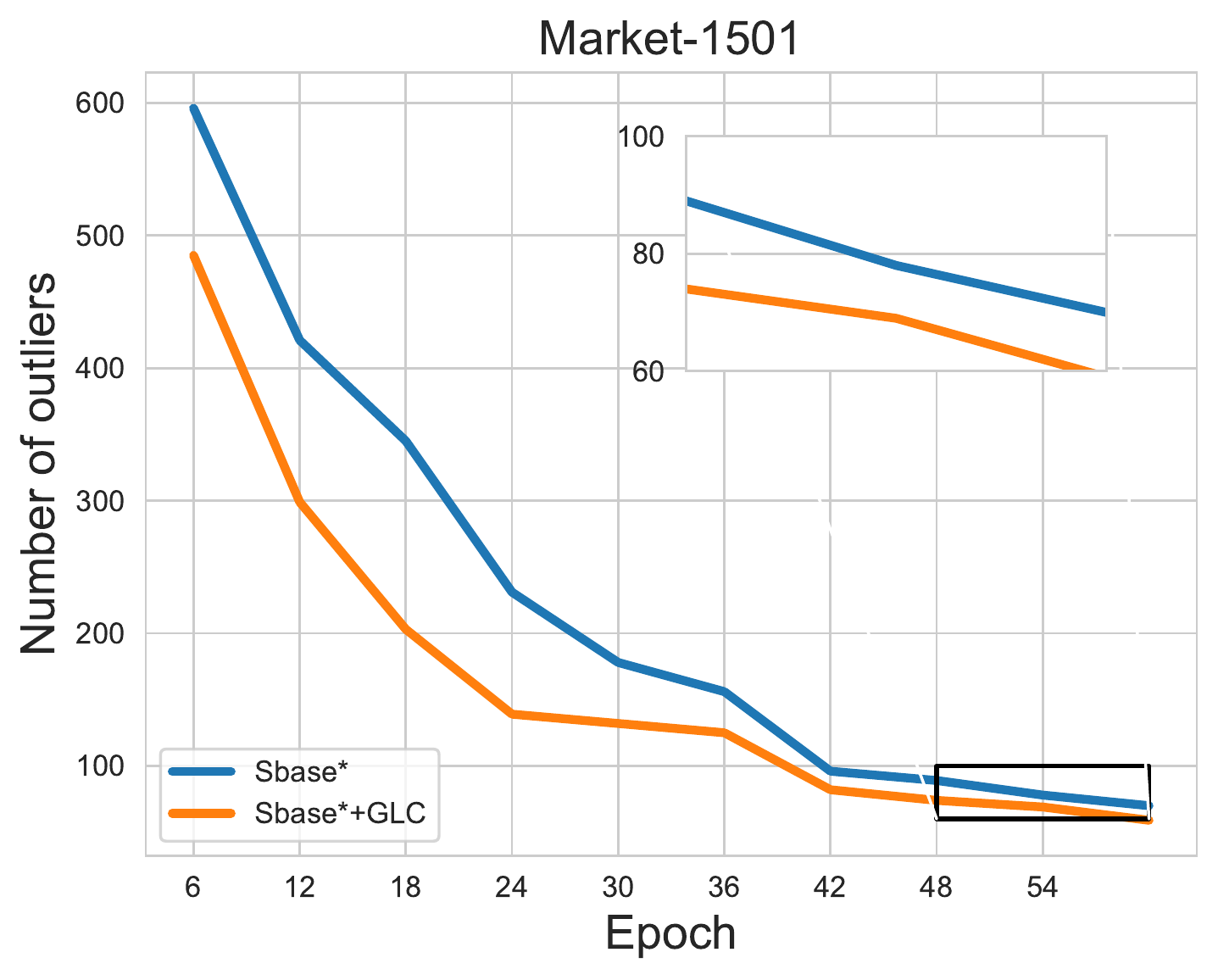}
    }
  \subfigure[CAP* and CAP* + GLC.]{
    \centering
  \includegraphics[width=0.4\linewidth]{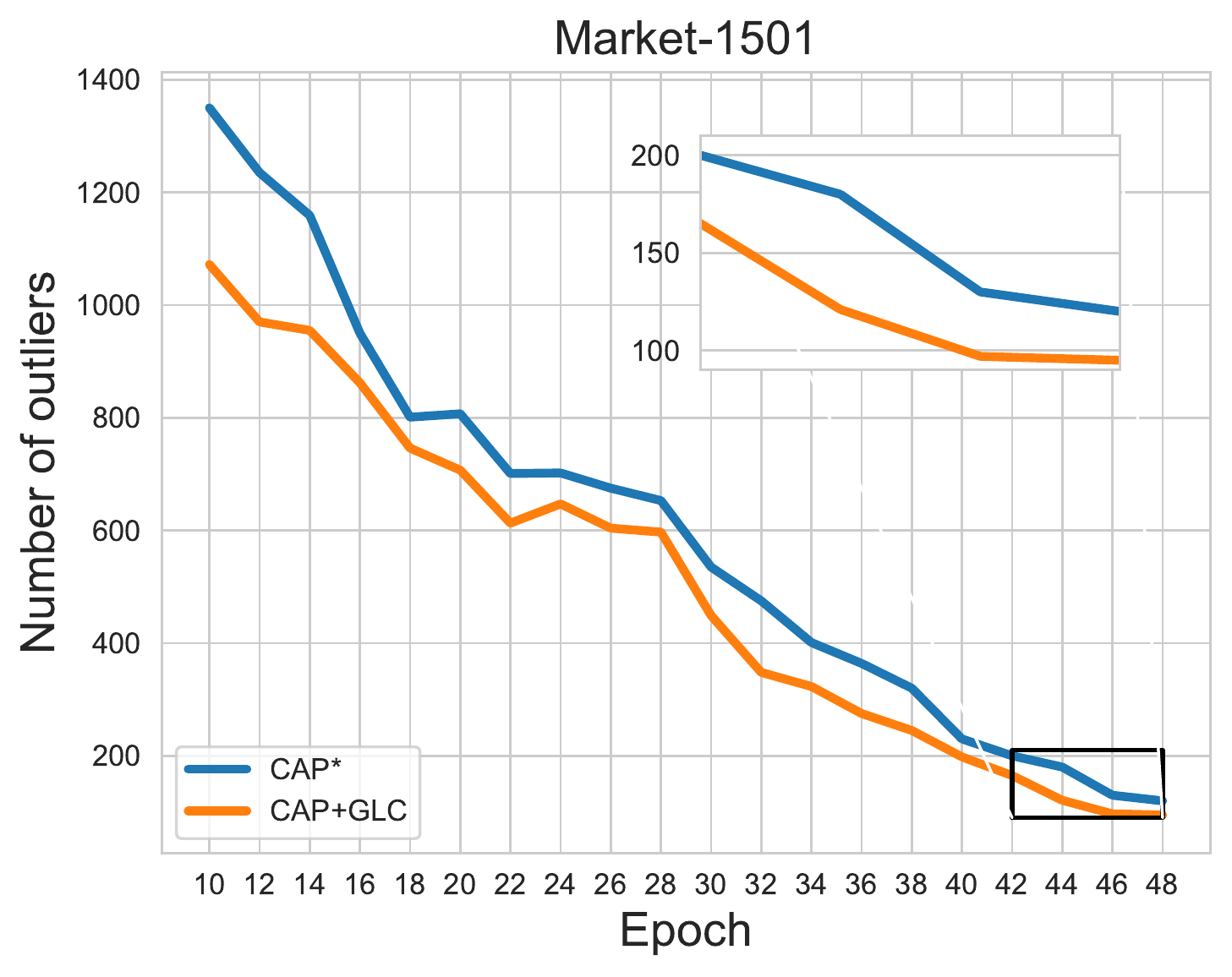}
  }
  \label{fig:ourliers}  
  \vspace{-1em}
  \caption{\textbf{(a-b)} NMI scores of pseudo labels at different epochs on Market-1501. Adding GLC consistently improves the pseudo labels. \textbf{(c-d)} Number of outliers in pseudo labels at different epochs on Market-1501. Adding GLC can consistently reduce the outliers.}
 \vspace{-1em}
\label{fig:nmi}
\end{figure}

\textbf{Effectiveness of label correction network}.
From \cref{tab:ablation_u} and \cref{tab:ablation_uda}, we can see that GLC effectively improves the performance of baseline methods on all benchmarks. 'Sbase* + GLC' will lead to 1.5\% and 1.8\% mAP improvement on MSMT-to-Market and Market-to-MSMT benchmarks while the mAP of CAP is promoted by 0.9\% and 1.1\% on Market and MSMT datasets with GLC.
The quality of pseudo labels largely affects the optimization of the feature extraction network.
In \cref{fig:nmi}, we illustrate corresponding Normalized Mutual Information (NMI) and number of outliers of the pseudo labels at different epochs with/without our GLC. 
It can be clearly observed that, by using GLC, the quality of pseudo labels is improved and the number of outliers is effectively reduced.

\begin{figure}[h]
  \centering
 \subfigure{
 \centering
 \includegraphics[width=0.4\linewidth, ]{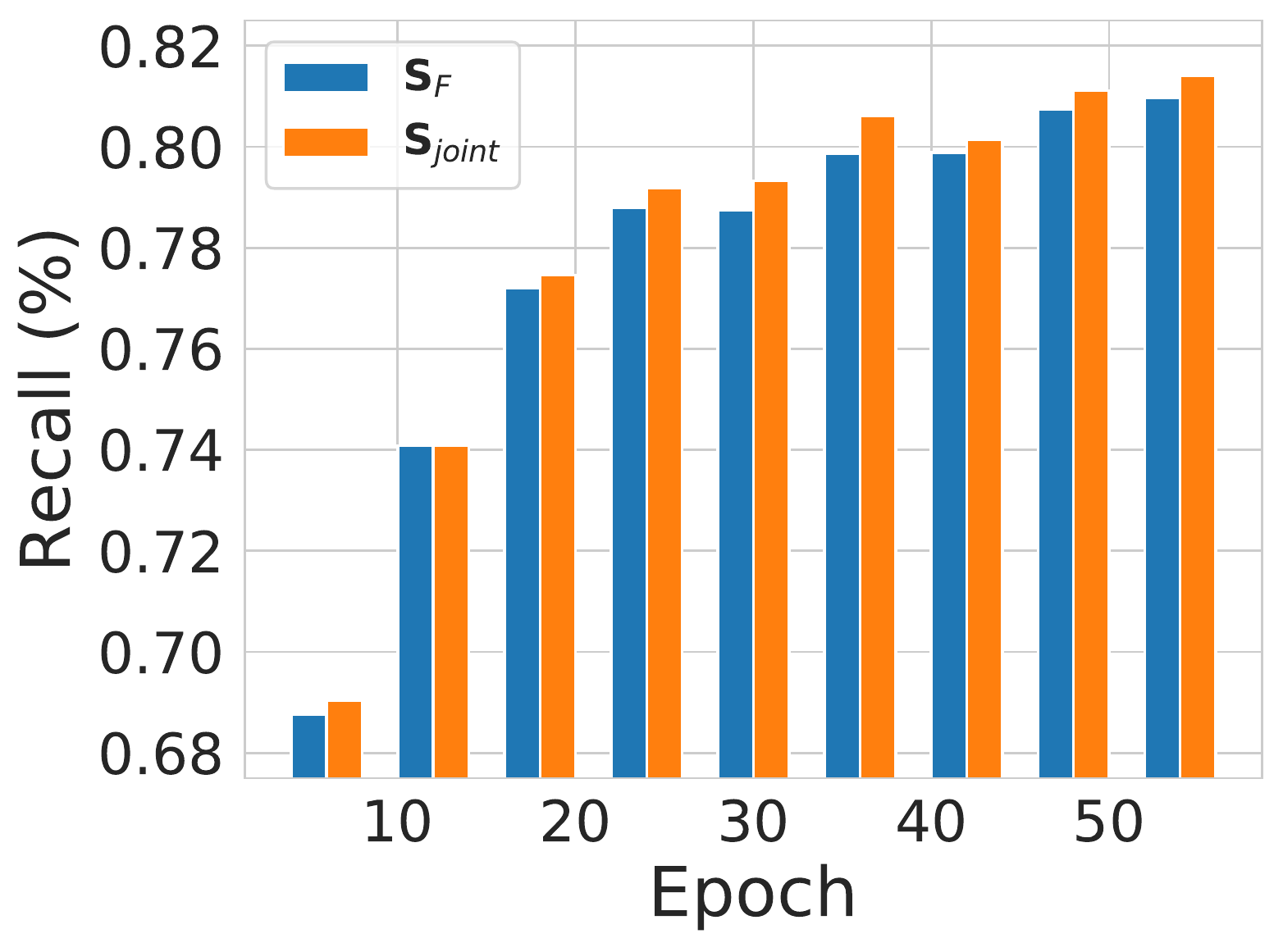}
 }\hspace{1em}
\subfigure{
\centering
\includegraphics[width=0.4\linewidth]{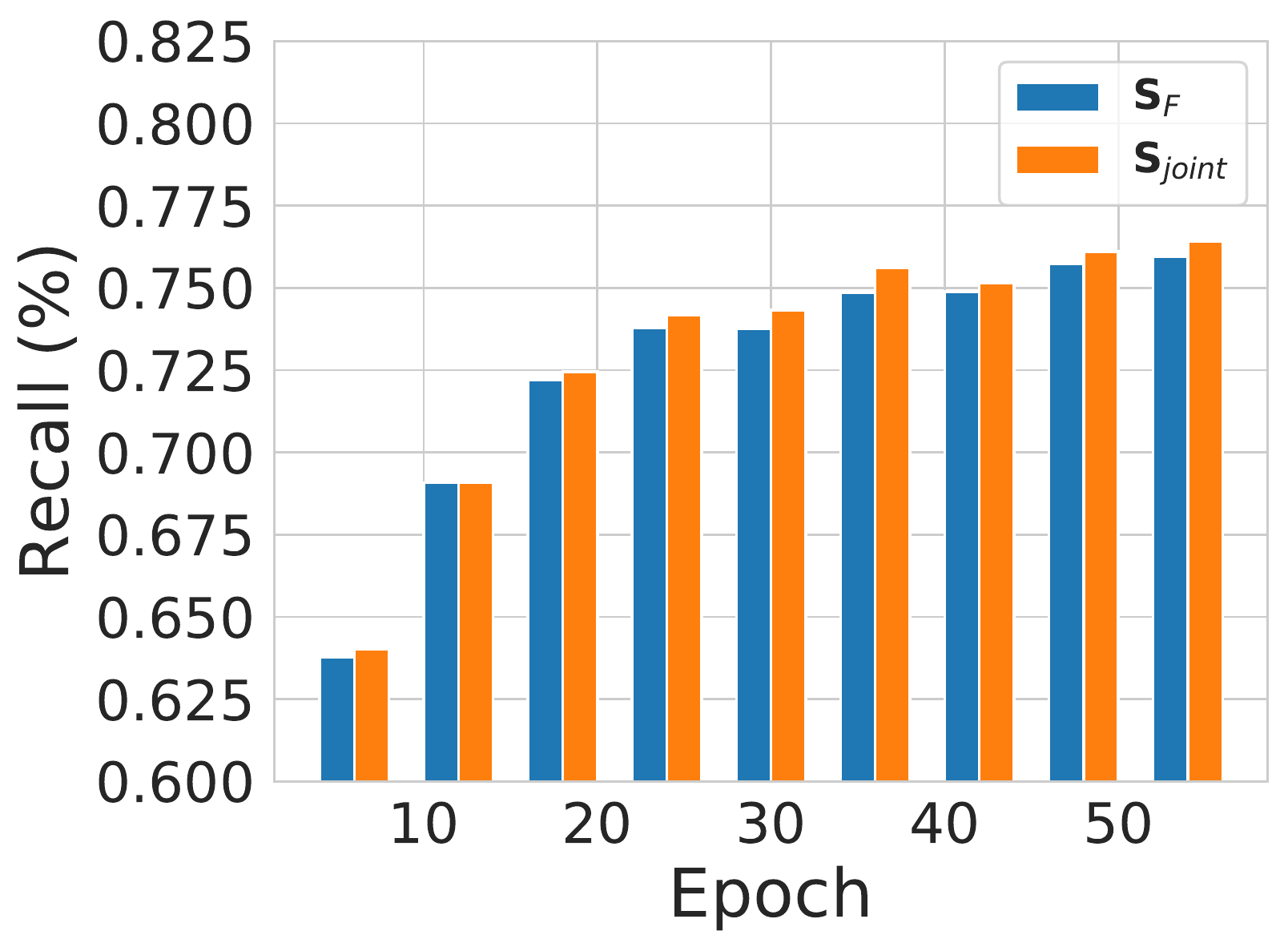}
}
  \caption{Recall of the kNN graph on Market1501 and MSMT17 at different epochs. $\mathbf{S}_F$ is the similarity by only using features. $\mathbf{S}_{joint}$ is the similarity by jointly using features and classification scores.}
  \label{fig:knn recall}
\end{figure}

\textbf{Effectiveness of joint similarity measurement}.
In order to connect more images of the identical person together in the $k$NN graph construction, we propose to jointly combine features and classification scores to obtain more robust sample similarity measuerment. In \cref{tab:ablation_u} and \cref{tab:ablation_uda}, we can see that GLC outperforms GLC$^-$ consistently. We also compute the recall of the constructed $k$NN graph to illustrate the effectiveness of joint similarity more intuitively. From \cref{fig:knn recall} we can see that the $k$NN graph constructed with joint similarity consistently obtains higher recall than that with just feature similarity. Since the classification score is more robust to data distribution gap like camera viewpoint, background and occlusion, adopting joint similarity can help $k$NN graph involve more hard positive edges. For example, the $k$NN graph can associate images of the identical person from different cameras or in different occlusion situation together. With the improved $k$NN graph, GLC obtains more discriminative node features and thus refine the pseudo labels better.

\textbf{Effectiveness of restart training strategy}.
The restart training strategy makes sense from eliminating the accumulated errors inherited by network parameters. With the help of the pseudo labels of higher quality, the re-initialized feature extractor will be trained again with a better start. The results in \cref{tab:ablation_u} and \cref{tab:ablation_uda} demonstrate that the restart strategy can bring consistent performance improvement. \cref{fig:restart_nmi} illustrates the mAP and NMI variation trend of the baseline method Sbase with and without the restart strategy during the training. After restarting, the quality of labels maintains at a high level and can be improved after a short decline. The mAP experiences a short decline and soon climbs to a higher point. 

\begin{figure}[t]
  \centering
 \subfigure{
 \centering
 \includegraphics[width=0.4\linewidth, ]{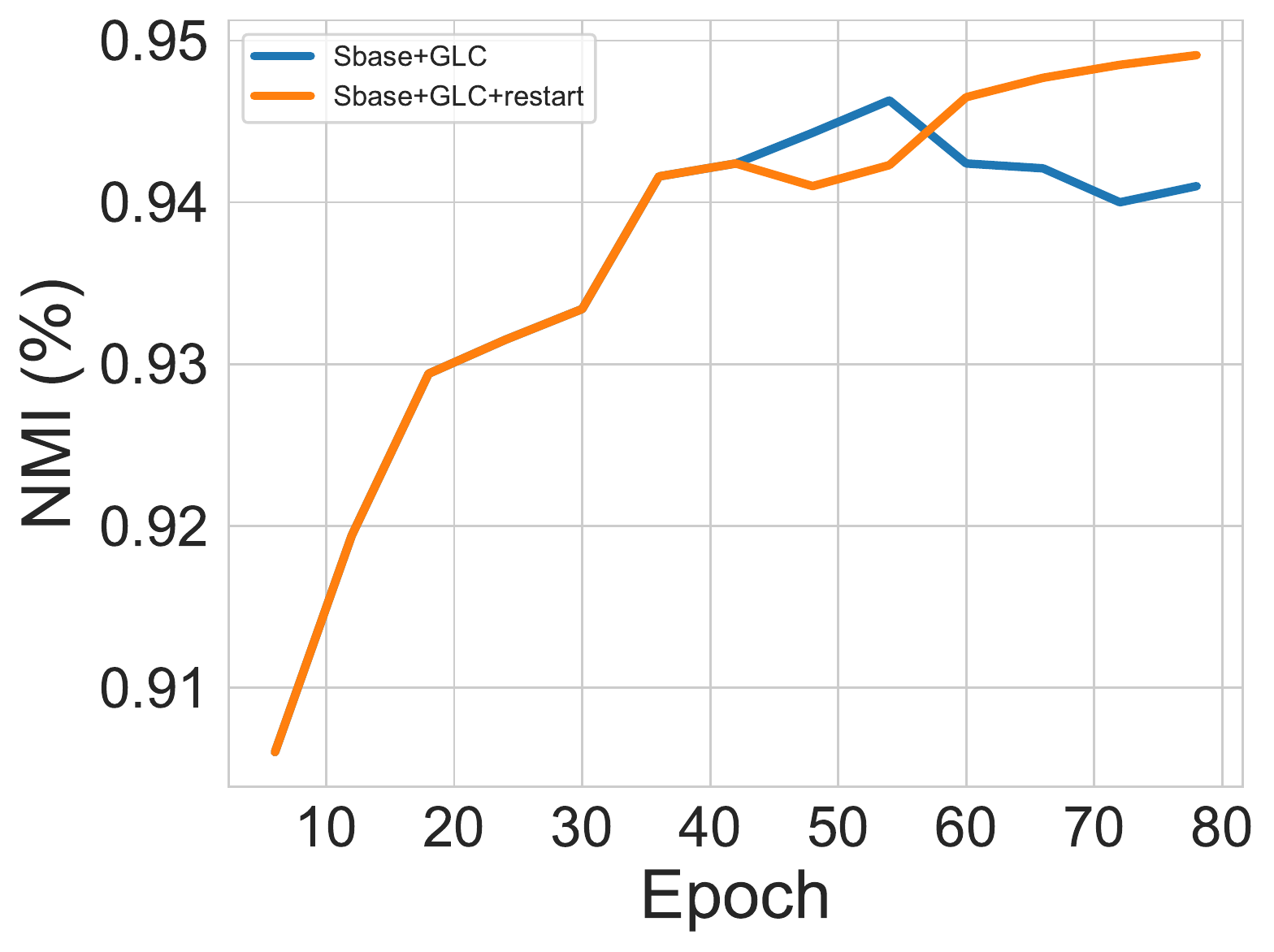}
 }
\hspace{1em}
\subfigure{
 \centering
\includegraphics[width=0.4\linewidth]{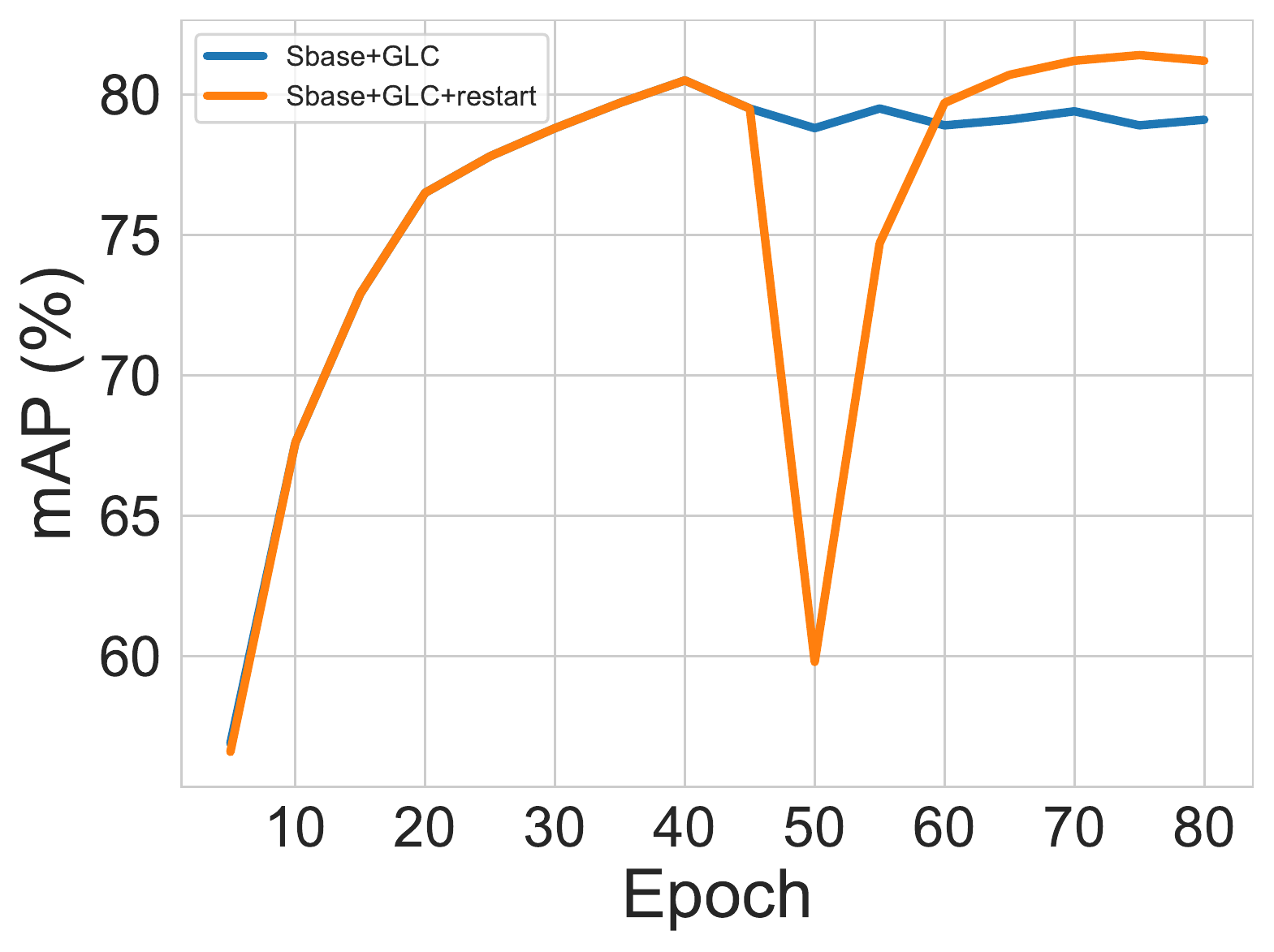}
}\vspace{-1em}
  \caption{NMI scores and mAP scores of Sbase*+GLC and Sbase*+GLC+restart on Market-1501.}
  \label{fig:restart_nmi}
\vspace{-1em}
\end{figure}

\subsection{Parameter Analysis}
\begin{figure}
\centering
  \subfigure{
    \centering
    \includegraphics[width=0.22\linewidth]{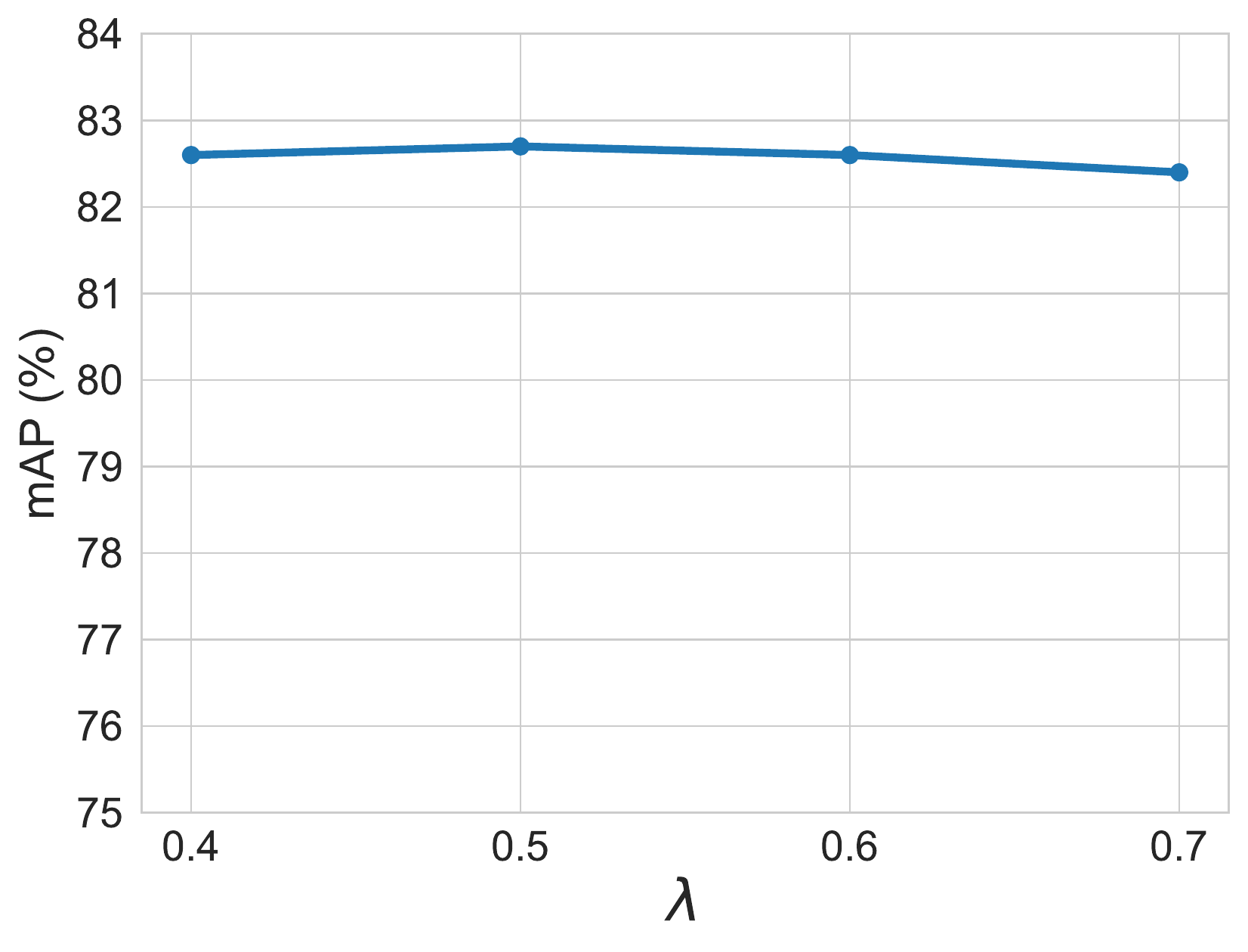}
    }
\subfigure{
    \centering
    \includegraphics[width=0.22\linewidth]{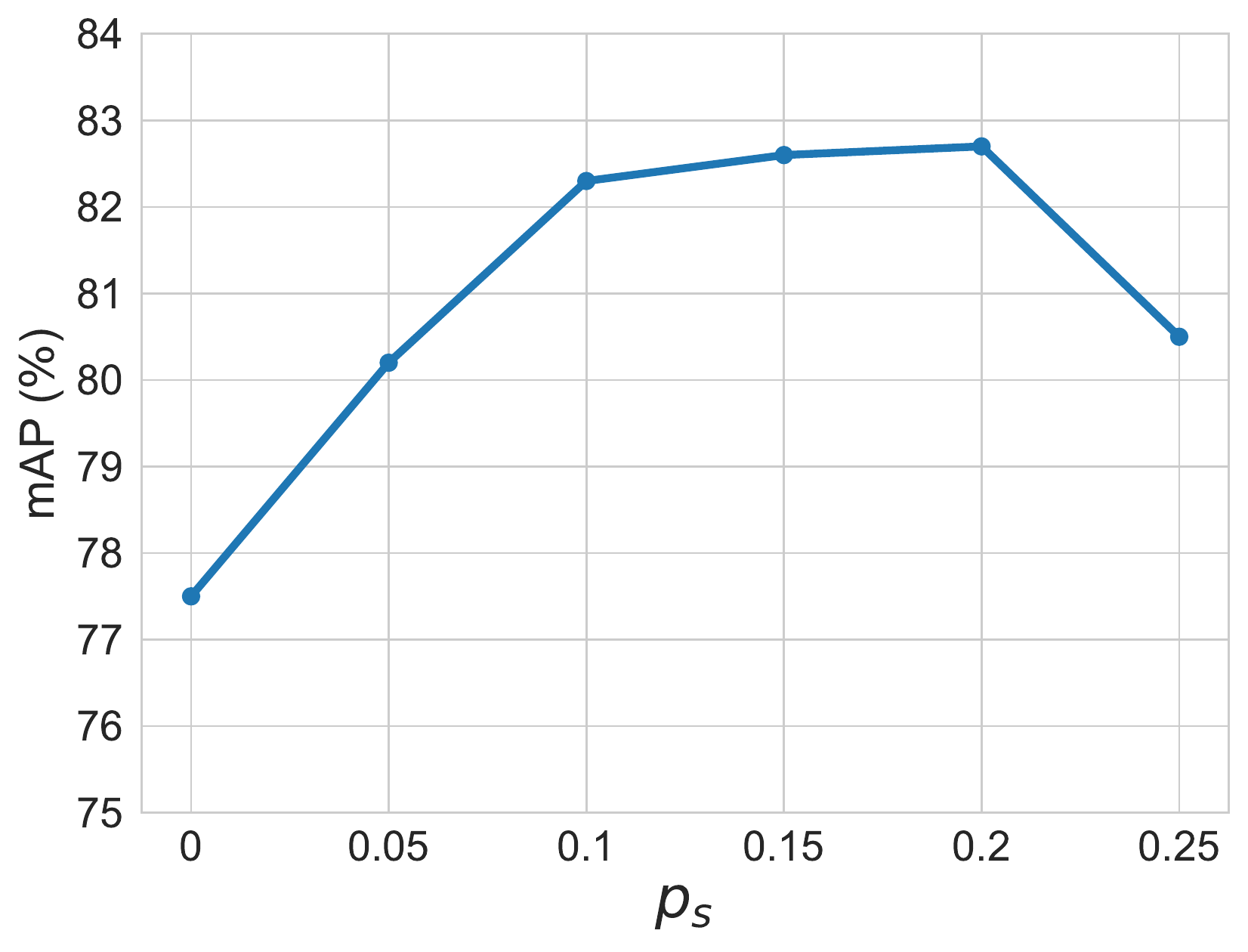} 
    }
\subfigure{
    \centering
    \includegraphics[width=0.22\linewidth]{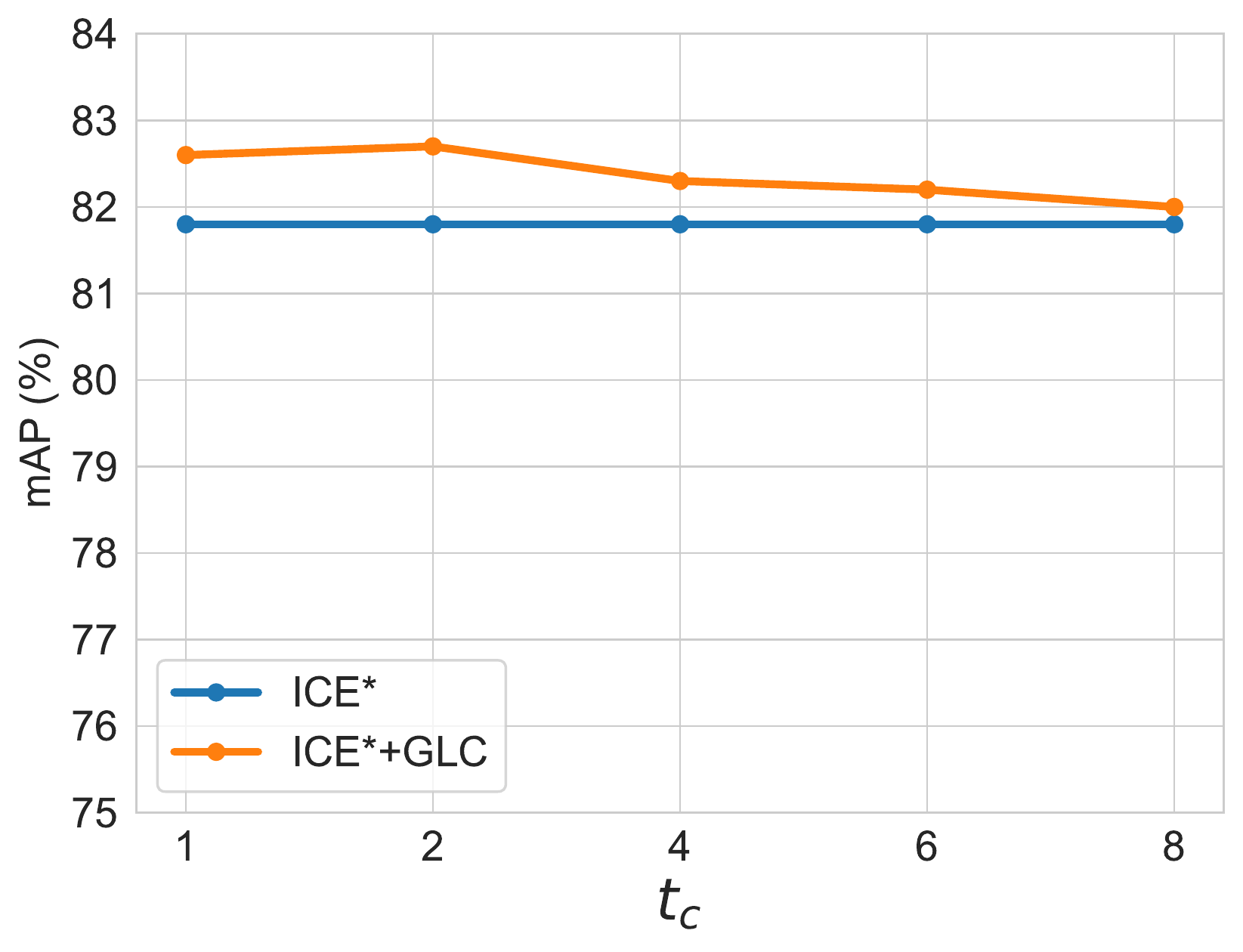}
     }
\subfigure{
    \centering
    \includegraphics[width=0.22\linewidth]{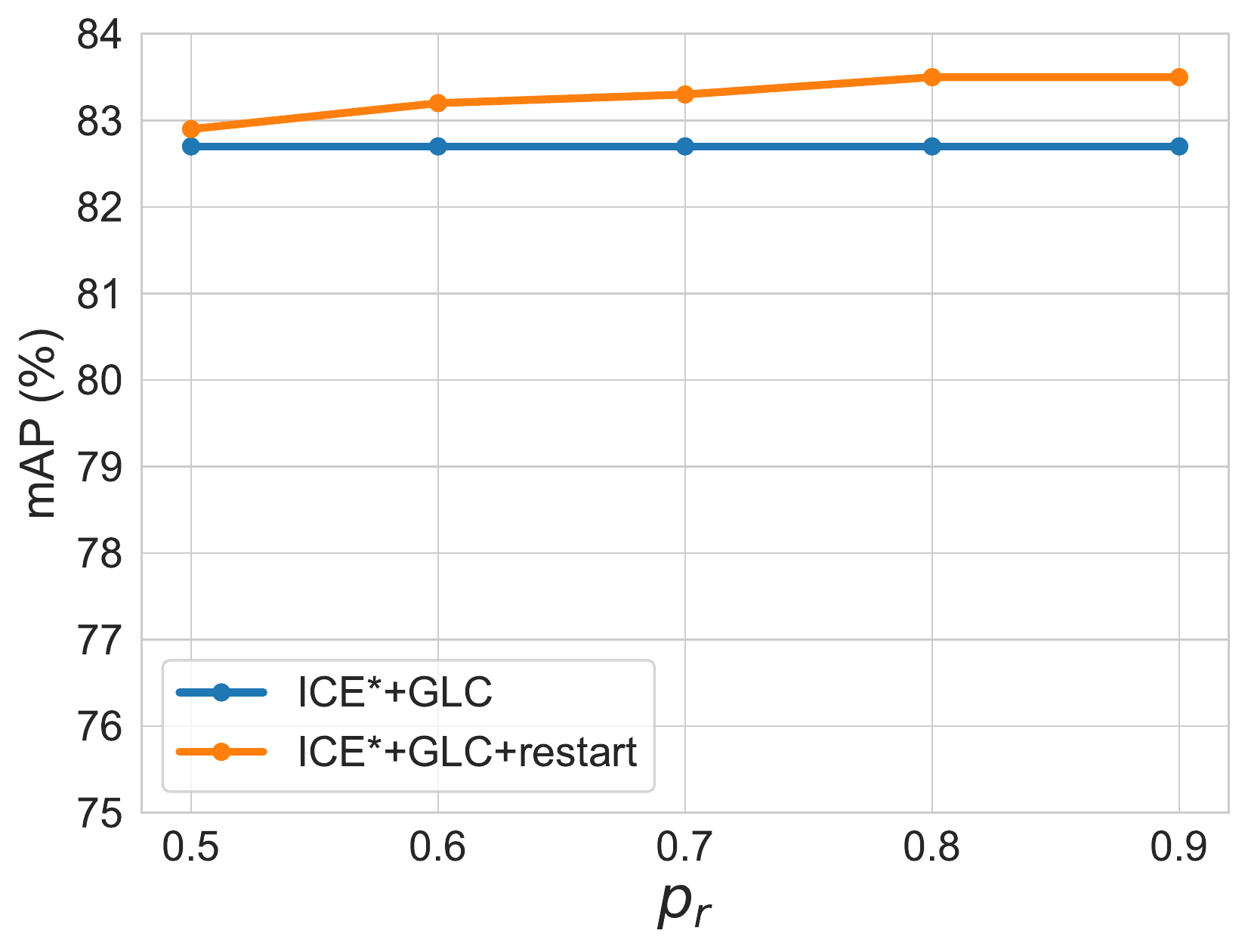}
     }
\vspace{-1em}
\caption{Parameter analysis on Market-1501.}
\label{fig:para}
\vspace{-1em}
\end{figure}

Our method involves several hyper-parameters: the weight $\lambda$ in joint similarity measurement, the ratio $p_s$ of the beginning epoch we adopt GLC to the whole training epochs, the interval $t_c$ between the consecutive two times we adopt GLC to rectify pseudo label, the ratio $p_r$ of the beginning epoch we re-initialize the feature extraction network to the whole training epochs, the early-stop timing $t_e$ in GLC training.
We conduct experiments on Market1501 with the baseline method ICE to analyze the above parameters and show the results in \cref{fig:para}. Larger $\lambda$ denotes we count more on raw features in similarity measurement.
Larger $p_s$ denotes we begin to rectify the noisy pseudo labels at a later training stage.
Larger $t_c$ means it takes longer to conduct the next label correction via GLC from the last time. Larger $p_r$ means that we restart the training of feature extraction network with parameter re-initialization at a later training stage.

From \cref{fig:para} we can see that our method is insensitive to the value of $\lambda$. Thus we set $\lambda$ to 0.5 in all our experiments.
As for when to add the GLC, we observe that adding GLC too early or too late in the training process is sub-optimal. 
On the one hand, the features are not discriminative enough at the early training stage, leading to unreliable initial pseudo labels and $k$NN graph. Thus the correction effect of GLC on the initial pseudo labels is limited. On the other hand, adding GLC too late means there are already many errors accumulated in the features and pseudo labels, leading to a limited correction effect of GLC. The best mAP is achieved when setting $p_s=0.2$. If the whole training epoch number is 50, $p_s=0.2$ means we start to adopt GLC to correct the pseudo labels after clustering at $0.2 \times 50 = 10$-th epoch. Setting smaller $t_c$ means we will adopt GLC to correct the pseudo labels more frequently. As we can see, a smaller $t_c$ will lead to marginally higher performance. However, no matter how many times GLC is adopted during the training, the performance is consistently higher than the baseline. Adding GLC brings extra time consumption during training when conducting unsupervised person ReID experiments with ICE and CAP, which conduct clustering every epoch, we set $t_c=2$. When conducting UDA person ReID experiments with Sbase and IDM, which conduct clustering every 6 epochs, we set $t_c=6$. The timing to add the restart strategy is also important. It is shown that restarting the feature extraction network later leads to a higher ReID performance. Since reinitializing the network from scratch in the middle of the network training will postpone the network convergence, we set $p_r=0.8$ for a better tradeoff. As for the early-stop timing in GLC training, we find that $t_e$ is insensitive to datasets and we set $t_e=250$. 
\begin{figure}[t]
  \centering
 \subfigure{
 \centering
 \includegraphics[width=0.4\linewidth]{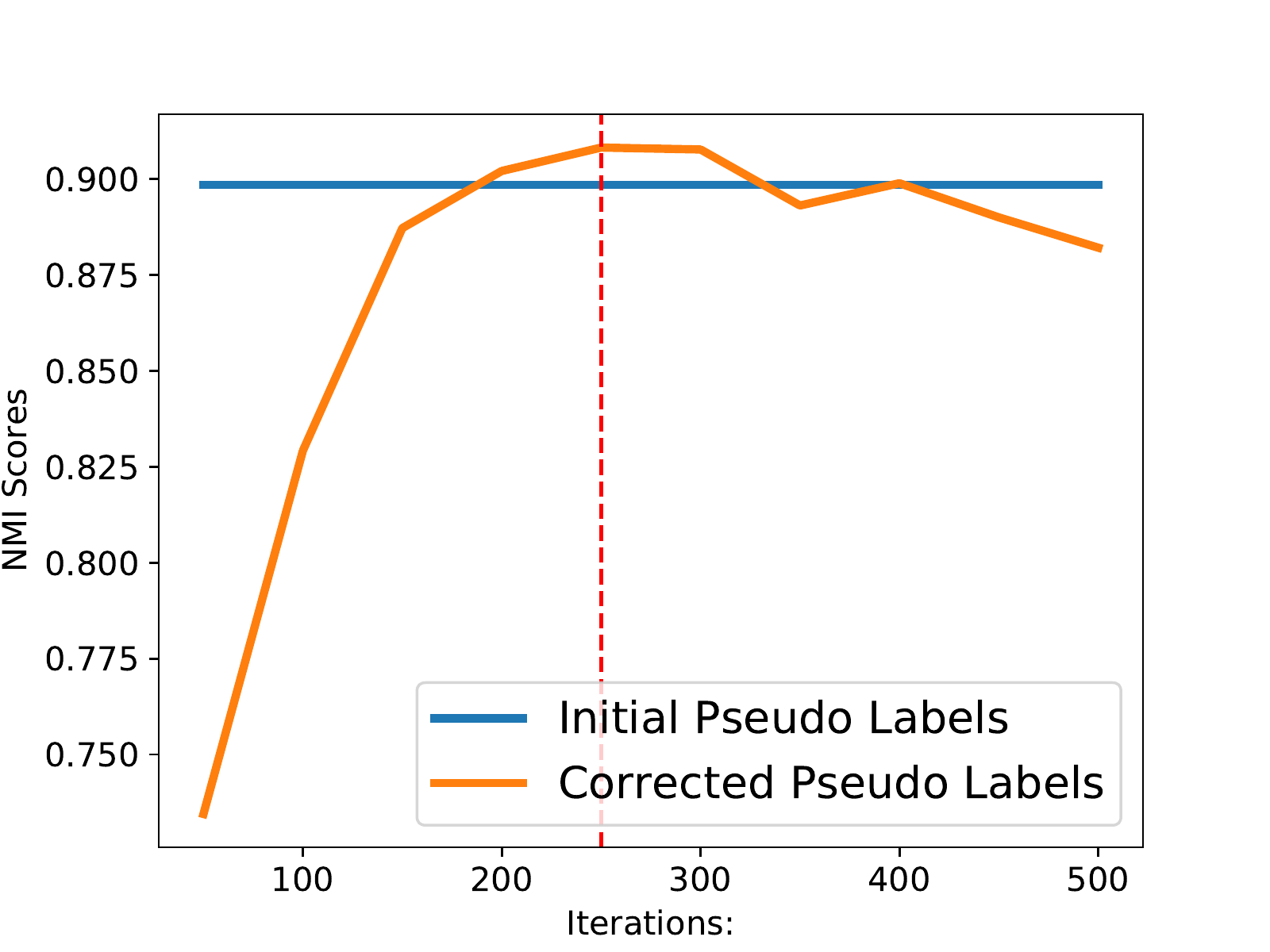}
 }
\hspace{1em}
\subfigure{
\includegraphics[width=0.4\linewidth]{NMI_Scores_in_GLC.pdf}
}
\vspace{-1em}
  \caption{Early stop at the certain timing $t_e$, which improves the pseudo labels consistently on both Market-1501 and MSMT17.}
  \label{fig:restart_nmi}
\vspace{-1em}
\end{figure}
\subsection{Discussion on Limitations.}
We propose GLC to correct pseudo labels effectively, but it also brings additional time consumption during the training. Each time we adopt GLC to correct the pseudo labels on Market-1501 and MSMT-17 consumes 0.09h and 0.16h, respectively. Nevertheless, from \cref{fig:para} we can see that a limited times of pseudo label correction by GLC can also effectively promote the performance of the network. For example, combing our method with IDM on Market1501-to-MSMT17 benchmark leads to a significant performance gain of 3.5\% at mAP and a total of 25\% training time increase.

\section{Conclusion}
Noisy pseudo labels substantially hinder the training of the feature extraction network. To tackle this issue, this paper first proposes a plug-and-play pseudo label correction network, perceives the data distribution from training samples, and adapts better to the dynamic feature at each epoch.
It learns to refine the initial noisy pseudo labels with the regularization of data relationship on the graph and the early stop training strategy. Besides, GLC is widely compatible with various clustering-based methods. 
Experiments on multiple benchmarks validate the effectiveness of our method.



\clearpage
%
%
\bibliographystyle{splncs04}
\bibliography{egbib}

\section{Parameters Analysis}
\subsection{Discussion of thresholds.}
\begin{figure}
    \centering
    \includegraphics[width=0.8\linewidth]{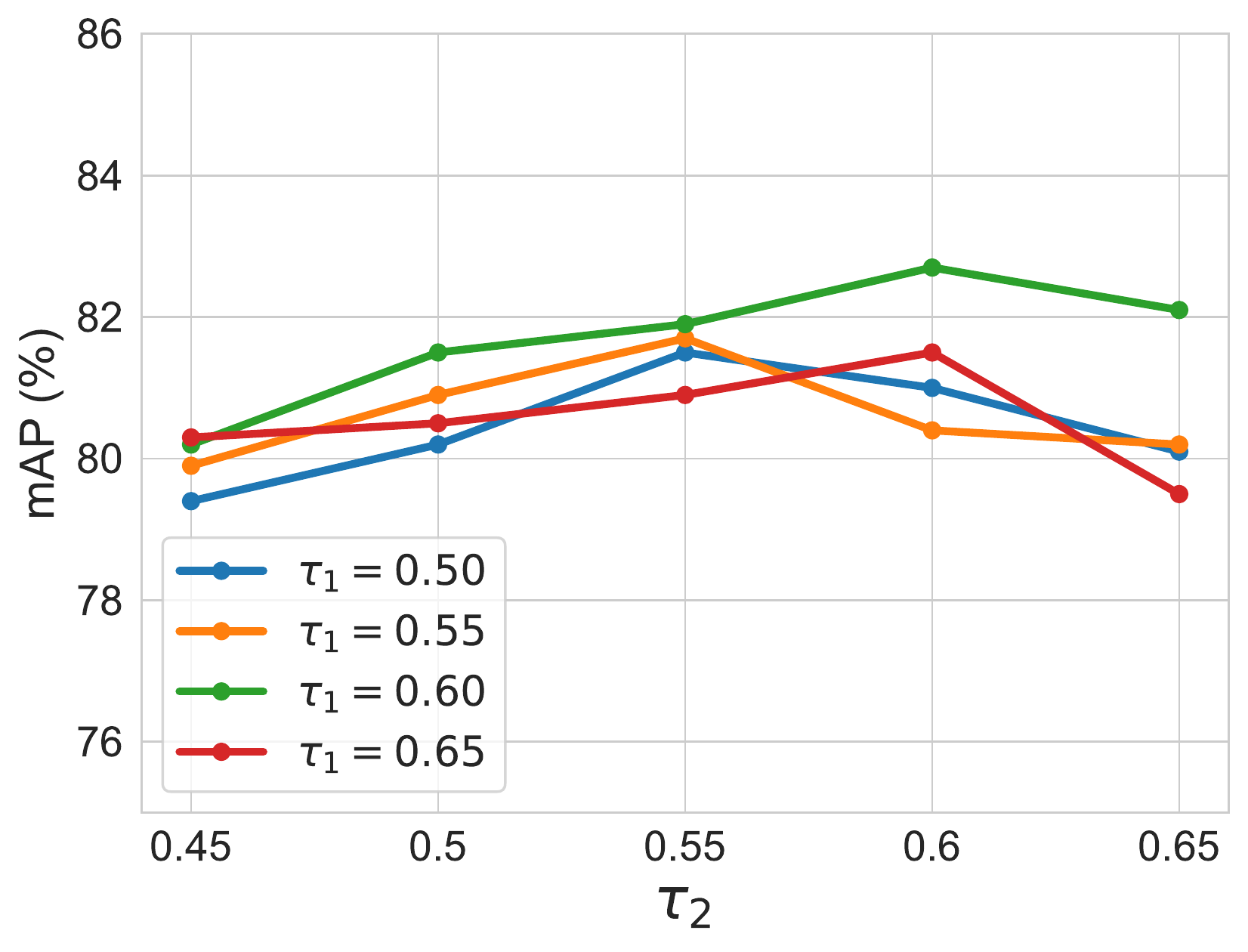}
    \caption{The mAP accuracy of ICE*+GLC with different $\tau_1$ and $\tau_2$ on Market1501.}
    \label{fig:t}
\vspace{-0.2cm}
\end{figure}

The ablation study of thresholds $\tau_1$ and $\tau_2$ in the inference of GLC is shown in \cref{fig:t}.
As we can observe, our method is not that sensitive to $\tau_1$ and $\tau_2$, and GLC with $\tau_1=\tau_2=0.6$ obtains the best performance.

\subsection{Discussion of the number of GCN layers.}
Too many GCN layers on the dense $k$NN graph may lead to the over-smoothing problem, so we follow the settings of GCN-based face clustering methods [6,24,25]. 
In our practice, more layers do not improve the quality of pseudo labels but even reduce the accuracy, as shown in \cref{tab:layer}.
\begin{table}[htbp]
  \centering
  \caption{NMI Scores when using the different number of GCN.}
    \begin{tabular}{c|cccc}
    \hline
    Layer Nums & 1     & 2     & 3     & 4 \\
    \hline
    Nmi Scores & 0.931 & 0.930  & 0.864 & 0.752 \\
    \hline
    \end{tabular}%
  \label{tab:layer}%
\end{table}%
\section{Additional Experimental Results on DukeMTMC-ReID}
Some additional experimental results on DukeMTMC-ReID, as shown in
 \cref{tab:uda} and \cref{tab:usl}. 
   \vspace{-2em}
\begin{table}[htbp]
  \centering
  \caption{Comparisons to the state-of-the-arts on DukeMTMC-ReID for unsupervised domain adaptive person ReID.}
    \begin{tabular}{c|ccc|ccc}
    \hline
    \multirow{2}[4]{*}{method} & \multicolumn{3}{c}{Duke to Market} & \multicolumn{3}{c}{Market to Duke} \\
\cline{2-7}          & mAP   & R1    & R5    & mAP   & R1    & R5 \\
    \hline
    SSG   & 58.3  & 80    & 90    & 53.4  & 73    & 80.6 \\
    MMT   & 73.8  & 89.5  & 96    & 62.3  & 76.3  & 87.7 \\
    SpCL  & 76.7  & 90.3  & 96.2  & 68.8  & 82.9  & 90.1 \\
    UNRN  & 78.1  & 91.9  & 96.1  & 69.1  & 82    & 90.7 \\
    OPLG  & 80    & 91.5  & -    & 70.1  & 82.2  & - \\
    TDRL  & 83.4  & 94.2  & -    & 70.8  & 83.5  & - \\
    \hline
    Sbase* & 78.8  & 91.9  & 96.9  & 68.6  & 81.2  & 90.7 \\
    Sbase+Ours$^-$ & 80.7  & 92    & 96    & 70    & 82.9  & 92.7 \\
    Sbase+Ours$^+$ & 82    & 92.4  & \textbf{97.4} & 71.1  & \textbf{84.3} & 91.6 \\
    UNRN* & 81    & 92.1  & 96.7  & 70.4  & 83.5  & 90.6 \\
    UNRN+Ours$^-$ & 82.1  & 92.6  & 97.5  & 71.4  & 83.3  & 91.2 \\
    UNRN+Ours$^+$ & \textbf{83.6} & \textbf{93.5} & 97.3  & \textbf{72.0} & 83.9  & \textbf{91.6} \\
    \hline
    \end{tabular}%
  \label{tab:uda}%
    \vspace{2em}
  \caption{Comparisons to the state-of-the-arts on DukeMTMC-ReID for unsupervised person ReID.}
    \begin{tabular}{c|c|cccc}
    \hline
    \multirow{2}[4]{*}{Method} & \multirow{2}[4]{*}{Reference} & \multicolumn{4}{c}{DukeMTMC} \\
\cline{3-6}          &       & mAP   & R1    & R5    & R10 \\
    \hline
    MMCL  & CVPR20 & 40.2  & 65.2  & 75.9  & 80 \\
    HCT   & CVPR20 & 50.7  & 69.6  & 83.4  & 87.4 \\
    SpCL  & NeurIPS20 & 65.3  & 81.2  & 90.3  & 92.2 \\
    OPLG  & ICCV21 & 65.6  & 79.8  & 88.6  & 91.6 \\
    CAP   & AAAI21 & 67.3  & 81.1  & 89.3  & 91.8 \\
    ICE   & ICCV21 & 69.9  & 83.3  & 91.5  & 94.1 \\
    \hline
    CAP*  & AAAI21 & 66.9  & 80    & 89.5  & 91.6 \\
    CAP+Ours$^-$ & this paper & 67.5  & 81.2  & 90.1  & 92.5 \\
    CAP+Ours$^+$ & this paper & 68.2  & 82.1  & 90.2  & 92.9 \\
    ICE*  & ICCV21 & 69.4  & 83    & 91.1  & 93.3 \\
    ICE*+Ours$^-$ & this paper & 70    & 83.3  & 91.4  & 93.7 \\
    ICE*+Ours$^+$ & this paper & \textbf{71.3} & \textbf{84.0} & \textbf{91.6} & \textbf{93.8} \\
    \hline
    \end{tabular}%
  \label{tab:usl}%
\end{table}%

\end{document}